\documentclass[journal]{IEEEtran}
\usepackage[utf8]{inputenc}

\usepackage{flushend}
\usepackage{epsfig}
\usepackage{epstopdf}
\AppendGraphicsExtensions{.tiff}

\usepackage{graphbox} 

\usepackage[cmex10]{amsmath}
\usepackage{amssymb}
\usepackage{multirow}
\usepackage{rotating}
\usepackage[table]{xcolor}
\usepackage{threeparttable}
\usepackage{url}
\usepackage{color,soul}
\usepackage{algorithmic}
\usepackage{algorithm}
\usepackage{verbatim} 
\newcommand{\etal}{\textit{et al.}}

\makeatletter
\newcommand\notsotiny{\@setfontsize\notsotiny{6.31}{7.18}} 
\makeatother

\begin{document}

\title{\textit{FedDropoutAvg}: Generalizable federated learning for histopathology image classification}%

\author{Gozde~N.~Gunesli*, Mohsin~Bilal, Shan~E~Ahmed~Raza, and Nasir~M.~Rajpoot,~\IEEEmembership{Member,~IEEE}

\IEEEcompsocitemizethanks{\IEEEcompsocthanksitem 
This work has been submitted to the IEEE for possible publication. Copyright may be transferred without notice, after which this version may no longer be accessible.

\em{Asterisk indicates corresponding author.}
}

\thanks{*G. N. Gunesli is with the Department of Computer Science, University of Warwick, UK (e-mail: Gozde.Gunesli-Noyan@warwick.ac.uk).}

\thanks{ M. Bilal is with the Department of Computer Science, University of Warwick, UK (e-mail: Mohsin.Bilal@warwick.ac.uk).}

\thanks{ Shan E Ahmed Raza is with the Department of Computer Science, University of Warwick, UK (e-mail: Shan.Raza@warwick.ac.uk).}

\thanks{N. M. Rajpoot is with the Department of Computer Science, University of Warwick, UK (e-mail: N.M.Rajpoot@warwick.ac.uk)}

}

\maketitle

\begin{abstract}
Federated learning (FL) enables collaborative learning of a deep learning model without sharing the data of participating sites. FL in medical image analysis tasks is relatively new and open for enhancements. In this study, we propose \textit{FedDropoutAvg}, a new federated learning approach for training a generalizable model. The proposed method takes advantage of randomness, both in client selection and also in federated averaging process. We compare \textit{FedDropoutAvg} to several algorithms in an FL scenario for real-world multi-site histopathology image classification task. We show that with \textit{FedDropoutAvg}, the final model can achieve performance better than other FL approaches and closer to a classical deep learning model that requires all data to be shared for centralized training. We test the trained models on a large dataset consisting of 1.2 million image tiles from 21 different centers. To evaluate the generalization ability of the
proposed approach, we use held-out test sets from centers whose data was used in the FL and for unseen data from other independent centers whose data was not used in the federated training. We show that the proposed approach is more generalizable than other state-of-the-art federated training approaches. To the best of our knowledge, ours is the first study to use a randomized client and local model parameter selection procedure in a federated setting for a medical image analysis task.
\end{abstract}


\section{Introduction}
\label{sec:intro}

\begin{figure*}[htb]
\centering
\begin{minipage}[b]{0.8\linewidth}
  \centering
  \centerline{\includegraphics[width=1\linewidth]{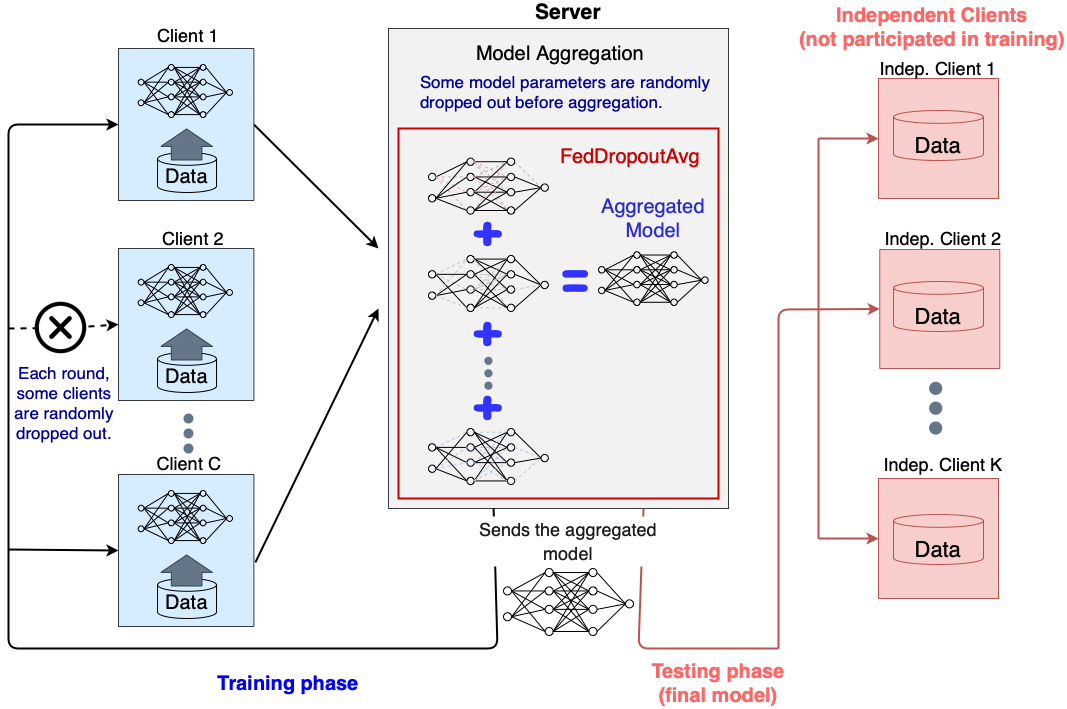}}
\end{minipage}
\caption{Workflow diagram of the proposed \textit{FedDropoutAvg} approach. \textbf{Training phase:} At the beginning of each training round, \textit{Central Server} sends global model weights to some clients which are randomly selected from all the clients participating in the training. Using the received model, local training takes place at each client on their local datasets. At the end of local training epochs, each \textit{Client i} sends the parameters of their locally trained model to \textit{Central Server}. Then, \textit{Central Server} randomly drops out some of the parameters of the received models and aggregates the models into a global model by averaging. This training process continues for some number of rounds. \textbf{Testing phase:} After the federated training is over, \textit{Central Server} sends the final model to independent clients for use on their own test sets, simulating a real life scenario.}
\label{fig:workflow}
\end{figure*}

In recent years, deep learning methods have shown success in many different tasks including those in computational pathology~\cite{litjens2017survey}. A major drawback to these approaches is the need for large amounts of data to train the networks. This drawback is even more obstructive in the medical field, as medical data are difficult to access and their sharing may be subject to legal and ethical limitations.

Federated learning (FL)~\cite{konevcny2016federated,mcmahan2017communication} allows to overcome these challenges. While in traditional deep learning approaches, all data is required to be co-located in a central server where model training takes place, in an FL paradigm, each of the multiple decentralized centers holding local data can train the model on their own servers. By enabling training of the deep learning models collaboratively without exchanging the datasets, FL offers a solution to data ownership and governance issues~\cite{kaissis2020secure}. Existing FL methods comprise of several rounds of local training and federal aggregation steps. In each round of the federated training process, each data-holder trains a model for some number of epochs on their local dataset. The local data-holders then send their trained models to a central server for model aggregation. The aggregated model is sent back to the data-holders for further training rounds. 

Model aggregation is an important step for the overall performance. The most common method of model aggregation in existing FL studies is Federated Averaging (FedAvg)~\cite{mcmahan2017communication}. FedAvg is weighted averaging of local model parameters (the gradients) to obtain a global model at each round. The weights in this case are determined based on the number of training samples of each local data-holder. Li {\etal}~\cite{li2018federated} argued that local models are often substantially different from global models because of the heterogeneous and imbalanced nature of the datasets. Therefore, they proposed FedProx, which contained a proximal term in the loss function to restrict the effects of local training and prevent divergence from the global model parameters. 

Study of FL in the area of medical image analysis is relatively new. Existing FL approaches in the medical imaging domain have focused on specific tasks including: analysis of brain imaging data~\cite{li2019privacy,sheller2018multi,roy2019braintorrent,li2020multi,silva2020fed,sarhan2020fairness}, CT hemorrhage segmentation~\cite{remedios2020federated}, breast density classification in mammography data~\cite{roth2020federated}, pancreas segmentation in abdominal CT images~\cite{wang2020automated} and classification of histopathology images~\cite{lu2020federated,andreux2020siloed}. Most of the FL studies in medical imaging have employed FedAvg method for model aggregation~\cite{li2019privacy,sheller2018multi,roy2019braintorrent,li2020multi,silva2020fed,sarhan2020fairness,roth2020federated, wang2020automated,lu2020federated}. Andreux \etal~\cite{andreux2020siloed} proposed an enhancement for aggregation of batch normalization (BN) layers. Remedios {\em et al.}~\cite{remedios2020federated} incorporated momentum in the gradient averaging method.

Medical image datasets can be very heterogeneous and unbalanced. Besides being unbalanced in terms of number of samples, the data quality and the diversity of different samples could differ by a large margin. In this case, the approach of FedAvg and FedProx, weighting the contributions of each local model by their data size, may have limitations. Since we cannot know beforehand which private local dataset(s) will generalize better for the test set of another center, measuring the contributions of individual centers and accurately weighting them is not feasible. While local validation set performances could be used to increase the weight of under-performing centers during training, these under-performing centers could have low data-quality and may not be worth learning strongly for the greater good. Therefore, we propose introducing dropout strategies for global model aggregation and clients participating in FL training to mitigate the complexities of learning from imbalanced and heterogeneous datasets from various clients.

In this paper, we propose Federated Dropout Averaging (FedDropoutAvg), a new FL aggregation method with the objective of obtaining a global federal model that achieves maximum performance by adjusting dropping out of the parameters of locally trained models before aggregation and also randomly dropping out some clients at each round. Our approach is inspired by FL model training with sensitive user data on mobile devices~\cite{konevcny2016federated,mcmahan2017communication}, where thousands of clients participate and several clients may also get dropped out at each training round due to various reasons and constraints like unstable connectivity or efficiency~\cite{mcmahan2017communication,li2018federated,li2019convergence,bonawitz2019towards,li2020federated}. However, the impact of clients dropping in FL training for medical image analysis tasks remained unexplored.

The proposed federated dropout averaging method for aggregating weights and parameters of deep neural network models trained at different client sites into a federated model is also different from the well-known dropout technique used to drop weights of the neurons during training of standard convolutional neural network by Srivastava \etal~\cite{srivastava2014dropout}. We present a comparative evaluation of the proposed method with locally trained models, the model trained in a centralized manner and two major FL model aggregation methods for histopathology image classification. We explore the comparative performance of the different methods on multi-institutional histopathology images datasets using federated models in real-world setting, using not only test data from centers (clients) participating in training but also the held-out test data from completely different centers.

\begin{figure*}[!ht]
\centering
\includegraphics[width=1\linewidth]{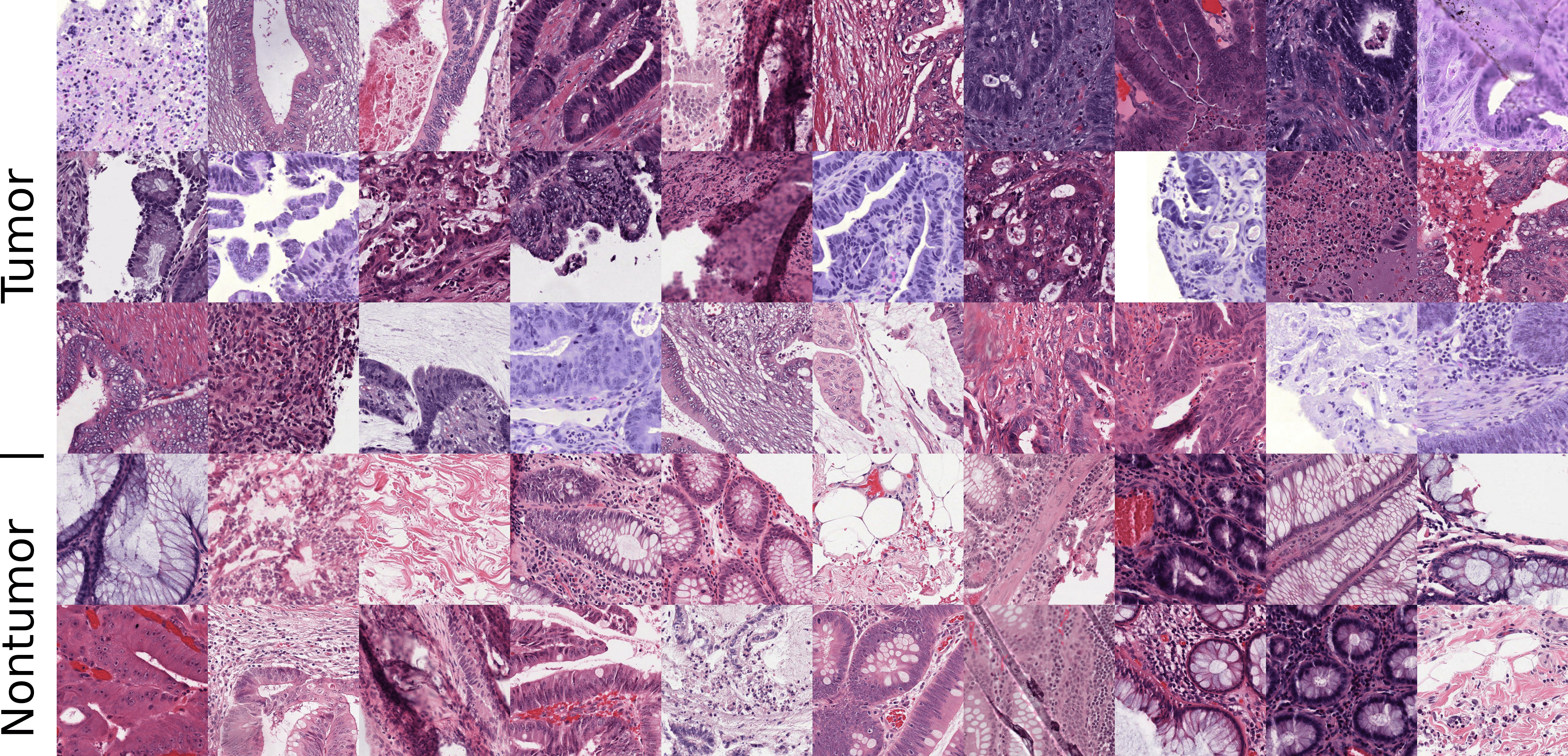}
\caption{Sample tumor (first three rows) and non-tumor (last two rows) image patches from different data centers. It can be observed that there is large intra-class variation between the two classes. Some image patches contain artifacts; stain color variation can also be seen in images from different centers.}
\label{fig:examples}
\end{figure*}

To be more specific, our work makes the following novel contributions:
\begin{itemize}
\item We present \textit{FedDropoutAvg}, a novel FL approach for federated training of deep learning models for histopathological image classification. 
\item We demonstrate the feasibility and superiority of our method using a large multi-site dataset. The data used in this study consists of 1.2 million images from 21 different centers. The individual datasets of these centers are imbalanced in terms of number of samples, patients, and number of positive and negative image patches and contain significant variation in the image data (color, brightness, focus) -- the so-called {\em domain shift}, as can be observed in Figure \ref{fig:examples} -- due to variations in scanning and staining parameters.
\item We show that our proposed approach outperforms previous federated strategies not only on the held-out test data of centers participated in training, but also on the data of independent centers whose data was not used in the training process at all.
\end{itemize}

\section{MATERIALS AND METHODS}
\label{sec:materialsandmethods}

In this section, we introduce the dataset and methods used in this study. An overall view of the proposed \textit{FedDropoutAvg} method can be seen in Fig.~\ref{fig:workflow}. 

\subsection{The Dataset}
\label{ssec:dataset}

\subsubsection{TCGA CRC-DX Dataset}
The dataset used in this study comprises of multi-gigapixel whole-slide images (WSIs) of 599 diagnostic slides from 591 colorectal cancer (CRC) patients contributed from 36 different centers to The Cancer Genome Atlas (TCGA) project. We used the Otsu thresholding~\cite{1979:otsu} to extract tissue region from each slide. The non-overlapping square tiles of size $512\times512$ were extracted from the segmented tissue region at $20\times$ magnification.
\subsubsection{Tumor Segmentation}
For tumor segmentation, we fine-tuned ResNet18~\cite{ResNet} pre-trained on ImageNet to distinguish between tumor and non-tumor tiles of each slide for the FL experiments. The total of 35436 tiles are extracted from seven randomly picked TCGA slides and two publicly available data sets~\cite{kather-msi,TIA-crc_grade}. Seventy percent of the data (24843 tiles) split for training, fifteen percent each for validation (5380  tiles), and held-out test set (5213  tiles with 2493 non-tumor tiles and 2720 tumor tiles). The network distinguished the tiles of the unseen test set belonging to the tumor and non-tumor classes with an accuracy of 99\%~\cite{bilal2021novel}. We used this trained network to separate the tumor and non-tumor tiles from the entire TCGA cohort. 

\subsubsection{Data Collection for FL Experiments}

Using the tiles and their labels as described above, we initially collected a multi-institutional dataset containing samples from 36 different centers. We have excluded centers contributing data of fewer than five patients from this study and randomly divided the dataset of the remaining 21 centers into federating training (11 centers, {\em Local Sets}) and independent test set (10 centers, {\em Independent Sets}). The data of {\em Local Sets} is patients-wise split into training, validation, and test sets ( $50\%$, $10\%$, $40\%$), keeping the tiles belonging to the same patient in the same set. Only the training set of {\em Local Sets} is used in model training and the validation set of {\em Local Sets} is used to select the best model parameters. In federated training, each local model is trained on the corresponding local training set of the clients in the {\em Local Sets}. In classical {\em Centralized} training, a single model is trained on the union of the training sets of the clients' data. Test sets of the clients in the {\em Local Sets} and all of the data of the clients in the {\em Independent Sets} have been used for evaluation purposes. More details are shown in Table~\ref{table:dataset}.

\begin{table*}[ht]
\centering
\caption{Number of patients ($n$), number of tumor ($N_t^i$) and non-tumor ($N_{nt}^i$) image patches (each patch being $512{\times}512$ pixels) per center (client $i$) and total number of patches ($N$) from the TCGA-CRC-DX dataset used in the training (TR), validation (VAL) and test (TS) sets. Training data from centers "CM" through "AA" used for training, their hold-out test sets and data from "QG" through "AD" used for testing. In the 3-fold experiments, at each fold, data from 7 different centers used for testing.} 

\scriptsize
\begin{center}
\tabcolsep=0.07cm
\renewcommand{\arraystretch}{1.5}
\begin{tabular}{||l|l||c|c|c|c|c|c|c|c|c|c|c||c|c|c|c|c|c|c|c|c|c||}
\hline
\multicolumn{2}{|l|}{\multirow{2}{*}{}}                           & \multicolumn{21}{|c|}{All Centers}                                                                                                                                       \\ \cline{3-23}
\multicolumn{2}{|c|}{}                                            & CM     & AY    & A6    & DY    & AF    & CK    & DC    & G4     & AG    & AH   & AA     & QG    & F4    & DM    & AZ    & CA    & D5    & NH    & EI    & F5    & AD   \\ \hline
\multirow{3}{*}{TR}  & \textbf{$n$} & 18     & 5     & 24    & 3     & 9     & 7     & 6     & 13     & 36    & 3    & 78     & 2     & 8     & 11    & 10    & 5     & 15    & 4     & 8     & 6     & 6    \\ \cline{2-23}
                     & \textbf{$N_{nt}^i$}        & 53.5K  & 4.7K  & 13.9K & 5.7K  & 6.9K  & 27.3K & 15.2K & 54.2K  & 8.5K  & 1.8K & 30.9K  & 12.3K & 23.7K & 24.0K & 31.0K & 8.5K  & 33.9K & 18.6K & 26.8K & 12.7K & 2.1K \\ \cline{2-23}
                     & \textbf{$N_t^i$}     & 27.0K  & 5.9K  & 11.9K & 5.5K  & 5.4K  & 9.2K  & 11.2K & 14.6K  & 19.4K & 1.0K & 46.0K  & 4.7K  & 12.0K & 7.1K  & 13.4K & 5.8K  & 9.6K  & 5.0K  & 7.1K  & 6.0K  & 2.5K \\ \hline
                     \hline
\multirow{3}{*}{VAL} & \textbf{$n$} & 4      & 2     & 5     & 1     & 2     & 2     & 2     & 3      & 8     & 1    & 16     & 1     & 2     & 3     & 3     & 2     & 4     & 1     & 2     & 2     & 2    \\ \cline{2-23}
                     & \textbf{$N_{nt}^i$}        & 14.1K  & 0.2K  & 1.8K  & 1.5K  & 1.2K  & 8.1K  & 4.0K  & 9.9K   & 1.6K  & 0.2K & 3.0K   & 2.3K  & 3.5K  & 6.9K  & 11.0K & 3.1K  & 3.1K  & 3.2K  & 4.0K  & 3.8K  & 299  \\ \cline{2-23}
                     & \textbf{$N_t^i$}     & 6.8K   & 1.9K  & 2.3K  & 1.3K  & 0.3K  & 4.2K  & 5.6K  & 3.5K   & 4.3K  & 1.0K & 11.4K  & 2.8K  & 3.2K  & 2.4K  & 3K    & 3.9K  & 1.7K  & 3.5K  & 399   & 3.5K  & 676  \\ \hline
                     \hline
\multirow{3}{*}{TS}  & \textbf{$n$} & 15     & 3     & 19    & 2     & 7     & 5     & 5     & 11     & 28    & 3    & 62     & 2     & 6     & 9     & 7     & 3     & 12    & 4     & 7     & 4     & 5    \\ \cline{2-23}
                     & \textbf{$N_{nt}^i$}        & 21.7K  & 0.4K  & 5.3K  & 2.6K  & 2.0K  & 12.0K & 12.2K & 37.2K  & 1.9K  & 0.3K & 4.0K   & 4.3K  & 7.1K  & 18.1K & 19.4K & 2.8K  & 5.9K  & 7.1K  & 6.9K  & 9.2K  & 653  \\ \cline{2-23}
                     & \textbf{$N_t^i$}     & 24.6K  & 4.1K  & 8.9K  & 1.4K  & 3.3K  & 8.6K  & 6.1K  & 16.6K  & 14.9K & 1.7K & 32.3K  & 1.6K  & 4.9K  & 6.1K  & 11.2K & 4.7K  & 3.5K  & 6.2K  & 3.8K  & 3.7K  & 2.8K \\ \hline
                     \hline
\multicolumn{2}{|c|}{\textbf{$N$}}       & 147.7K & 17.3K & 44.1K & 17.9K & 19.1K & 69.4K & 54.3K & 135.9K & 50.1K & 6.0K & 127.6K & 28.0K & 54.4K & 64.7K & 89.0K & 28.7K & 57.7K & 43.6K & 49.0K & 38.9K & 9.0K \\
\hline
\end{tabular}
\end{center}
\label{table:dataset}
\end{table*}

\subsection{Federated Dropout Averaging (\textit{FedDropoutAvg})}
\label{ssec:method}

\subsubsection{Dropping out model weights before aggregation}
\label{sssec:fdr}
Our proposed method is based on FedAvg proposed by~\cite{mcmahan2017communication} which is the most popular method used for model aggregation in federated systems. In FedAvg method, in each federated round $t$, global model parameters $\theta^{t+1}$  are calculated as follows:

\begin{equation}
\theta^{t+1} = \sum_{i=1}^{C} \alpha_i \theta_i^t
\end{equation}

Here, $\alpha_i \geq 0 $ is the contribution weight of each client $i$, while there are $C$ clients joining the federated training process.

In standard federated averaging, $\alpha_i$ is calculated as the proportion of number of data samples $N_i$ of each client $i$ to total number of samples of all clients participated in training $ N^{(t)}$.

\begin{equation}
\alpha_i = {\frac{N_i}{N}} \quad \textrm{where} \quad N = {\sum_{i=1}^{C} N_i}
\end{equation}

In the \textit{FedDropoutAvg} method, we propose to dropout some of the weights from each client model $\theta_i^t$ before aggregation and adjust the contribution weights accordingly. Here, we define a new parameter \textit{Federated Dropout Rate} ($fdr$), where $fdr = 0$ is same as the standard FedAvg. In our experimental settings, the best value for this parameter is selected as $fdr=0.3$ (when $cdr=0.2$). At the end of each round, we create random masks for each client model $\theta_i^t$, and use those masks to select the model parameters (weights) which will be included in the aggregation process.

For a more formal explanation, let parameter $p^t_{k,l,i}$ be any parameter (weights or biases) of any layer $l$ at index $k$ in model $\theta_i^t$. Then, $p^{t+1}_{k,l}$, the parameter at the same index of the aggregated global model will be calculated as follows:

\begin{equation}
p^{t+1}_{k,l} = \sum_{i=1}^{C} \alpha^t_{k,l,i} p^t_{k,l,i}
\end{equation}

where, $\alpha^t_{k,l,i}$ is the contribution weight of each client parameter $p^t_{k,l,i}$ and obtained as follows:

\begin{equation}
\alpha^t_{k,l,i} = {\frac{N_i R^t_{k,l,i}}{N^t_{k,l}}} \quad \textrm{where} \quad N^t_{k,l} = {\sum_{i=1}^{C} N_i R^t_{k,l,i}}
\end{equation}

Here, $N^t_{k,l}$ indicates the total number of data samples of all the clients whose parameter at layer $l$ at index $k$ are not dropped out from aggregation at the end of round $t$. 

The $R^t_{k,l,i}$ value in the calculation is a random Boolean value (0 or 1) which is obtained using the newly defined $fdr$. Where RandomUniform() draws a value from a uniform distribution over the half-open interval [0,1).:
\begin{equation}
R^t_{k,l,i} = (RandomUniform() > fdr)
\end{equation}

\subsubsection{Client dropout}
\label{sssec:cdr}
At each federated round, a random subset of clients are selected to participate in model training at that round. We have defined a new parameter \textit{Client Dropout Rate} ($cdr$) which modifies the the number of clients selected at each round. For a specific $cdr$, number of random clients selected at each round is constant. Please note that, for the case where $cdr = 0$, all of the clients will participate at each training round.

\subsection{Model}
\label{ssec:models}
We have used ResNet18~\cite{ResNet} model with group normalization (GN)~\cite{GN} layers instead of BN layers, as the GN layers are known to be more successful in decentralized machine learning settings~\cite{hsieh2020non}.

\section{Experiments and results}
\label{sec:resultsanddiscussion}
\begin{figure}[ht]
\centering
\includegraphics[height=7.2cm]{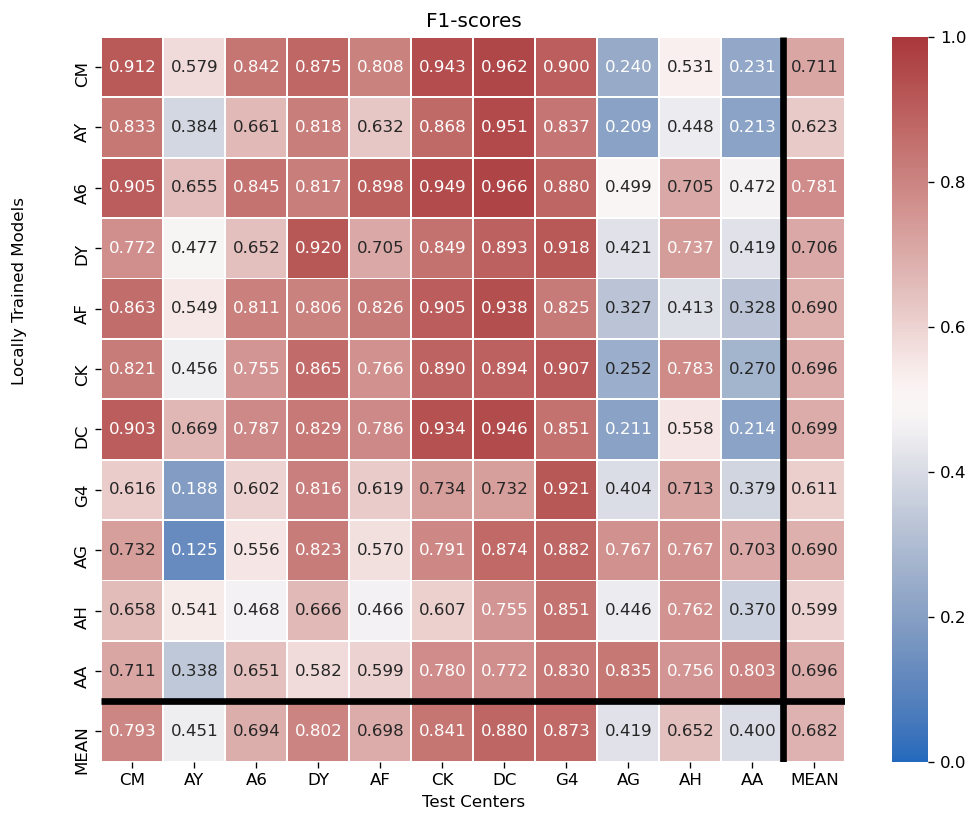}
\caption{Heatmap of F1 scores of the locally trained models (not federated) on the held-out test sets. Mean F1 of each locally trained model is given on the rightmost column.}
\label{fig:f1s-localmodels}
\end{figure}

\begin{figure}[ht]
\centering
\includegraphics[height=7.2cm]{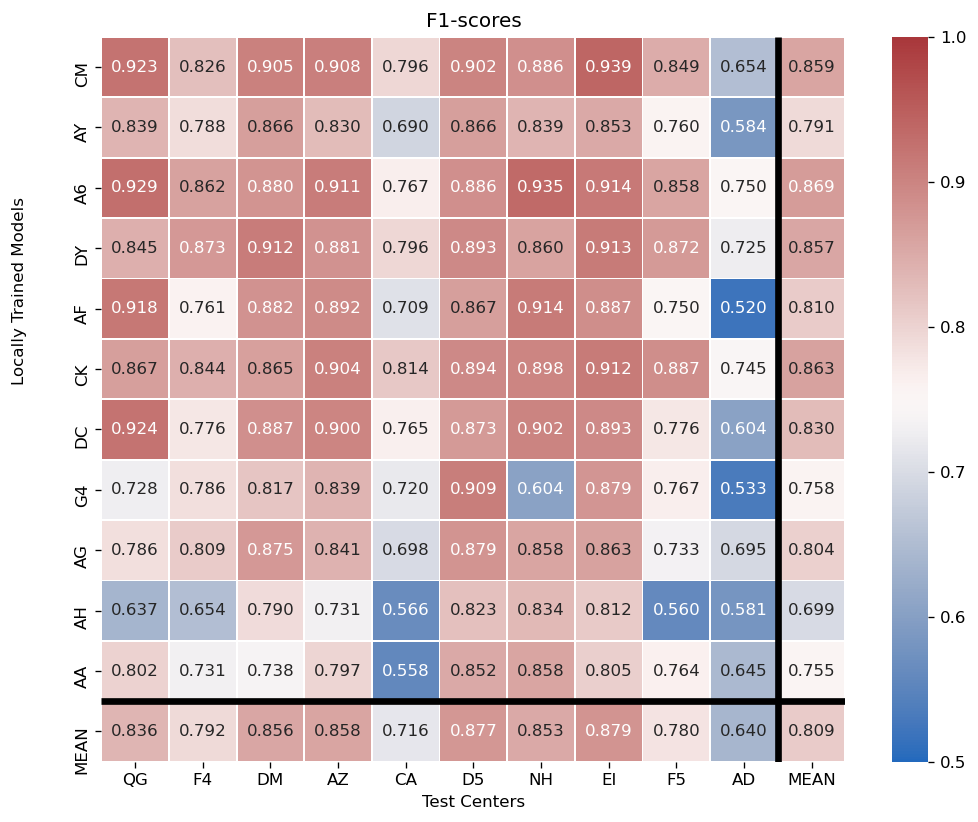}
\caption{Heatmap of F1 scores of the locally trained models (not federated) on the independent test centers. Mean F1 of each locally trained model is given on the rightmost column.}
\label{fig:f1s-localmodels-indep}
\end{figure}
\begin{figure}[ht]
\centering
\small{
\begin{tabular}{@{~}c@{~}}
\includegraphics[height=5.5cm]{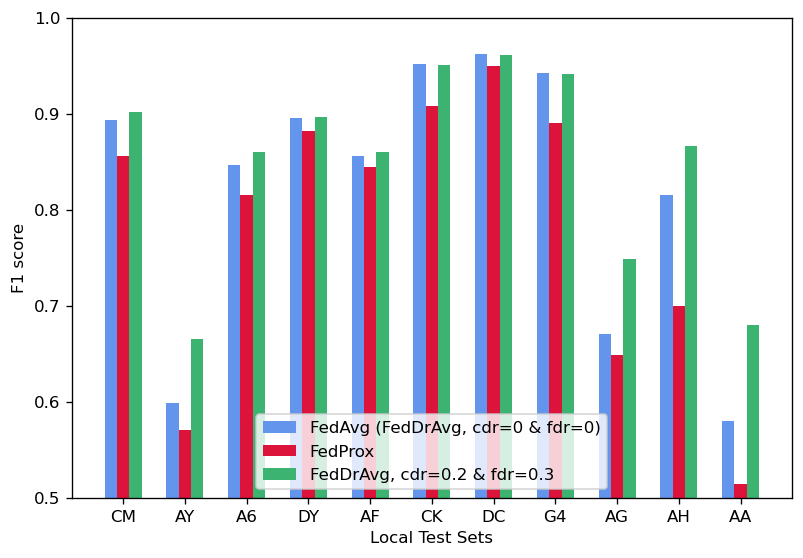} \\
(a) \\
\includegraphics[height=5.5cm]{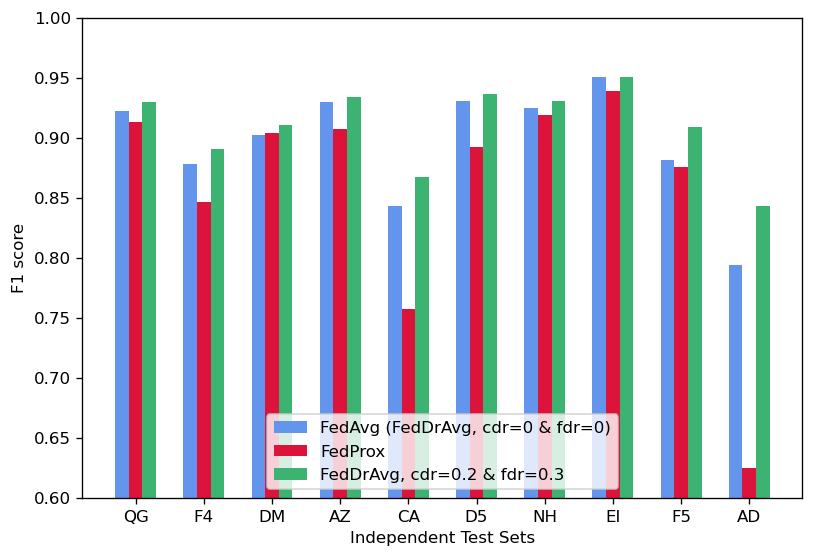} \\
(b) \\ 
\end{tabular}
}
\caption{F1 scores of the federated approaches on (a) the held-out test sets of the centers participated in training, (b) the datasets of the independent test centers (Note that these centers did not participate in the training of the models).}
\label{fig:f1-bars}
\end{figure}

\begin{figure}[ht]
\centering
\small{
\begin{tabular}{@{~}c@{~}}
\includegraphics[height=5.5cm]{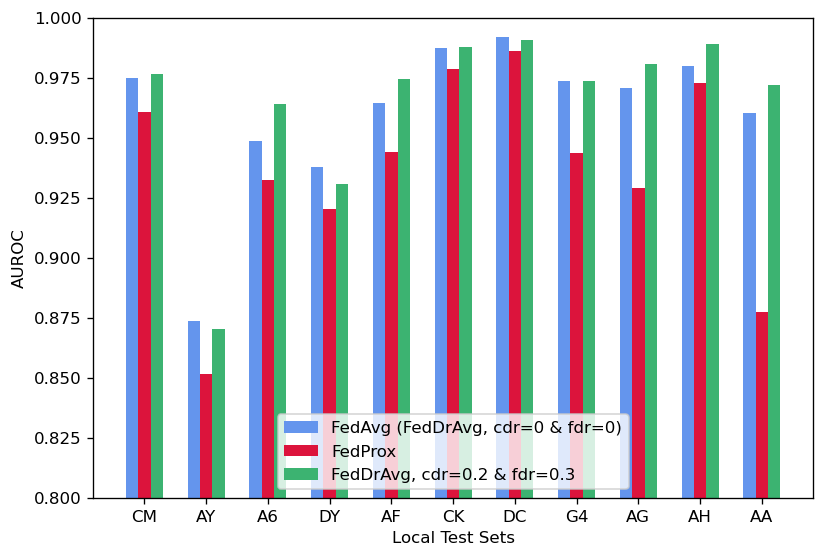} \\
(a) \\
\includegraphics[height=5.5cm]{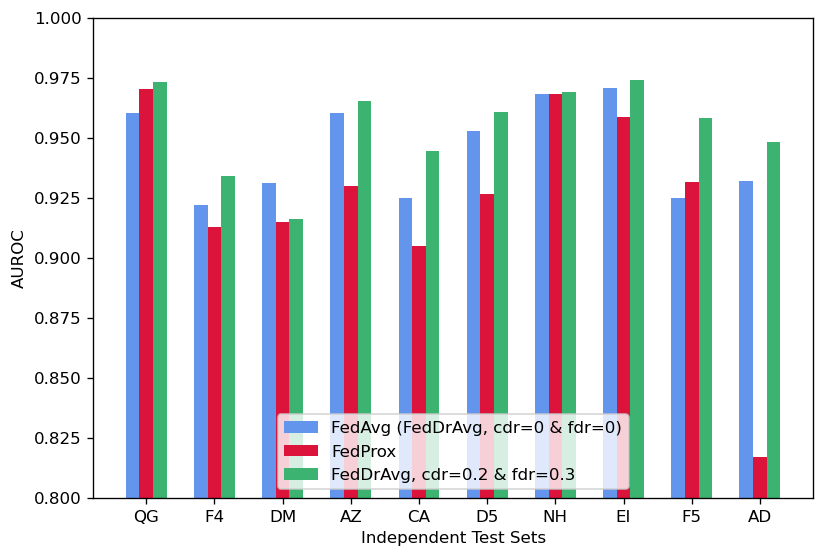} \\
(b) \\ 
\end{tabular}
}
\caption{Performance (AUROC) of the federated approaches on (a) the held-out test sets of the centers participated in training, (b) the datasets of the independent test centers (Note that these centers did not participate in the training of the models).}
\label{fig:auroc-bars}
\end{figure}

\subsection{Implementation and Training}
\label{ssec:implementation}
Local models and federated models are locally trained on the training data of clients in the {\em Local Sets}.  We have also trained the {\em centralized} model for comparison which is trained on data from all the training sets. For evaluation, we have compared the proposed \textit{FedDropoutAvg} method with \textit{FedAvg} and \textit{FedProx} methods.

All of the models are trained from scratch on GPU  for each comparison. For model training, we used class-weighted binary cross-entropy loss and SGD optimizer with initial learning rate 0.1, momentum 0.9 and weight decay 0.0001, with the learning rate halved after every 2 epochs. We have trained federated models (\textit{FedAvg}, \textit{FedProx} and proposed \textit{FedDropoutAvg}) for 20 rounds (one epoch per round) and selected the best model based on total cross-entropy loss on local validation sets. Similarly, local models and {\em centralized} model are trained for 20 epochs and models from the epochs with the best validation loss have been selected. 

\textit{FedAvg} and \textit{FedProx} models are trained in same settings with the proposed \textit{FedDropoutAvg} model. The proposed \textit{FedDropoutAvg} model is same with \textit{FedAvg} model when $cdr=0$ and $fdr=0$. For \textit{FedDropoutAvg} model, the best $cdr$ and $fdr$ parameters are selected on validation set as $0.2$ and $0.3$, based on a grid search on $cdr\in[0,0.1,0.2,0.4]$ and $fdr\in[0,0.1,0.2,0.3,0.4]$. \textit{FedProx} method has a parameter ($\mu$) which adjusts the effect of the proximal term on the loss function. This parameter is selected as 0.01 after a grid search from {0.5, 0.1, 0.01, 0.001} based on the performance on the validation set. For the implementation of the models and methods, we used PyTorch. 

\subsection{Experimental Results}
\label{ssec:results}
In this section, we present results of comparative analysis of our method with other FL methods, local training and centralized training approaches. We also analyze the effect of different {\em cdr} and {\em fdr} parameters on the performance.

\begin{table*}[ht]
\centering
\begin{center}
\caption{Average performance of the centralized and federated approaches on the held-out test sets of the centers participated in training and datasets of the Independent Centers (not participated in training). Performance given as the mean and standard deviation of the F1 scores and AUROC values. F1 scores and AUROC's on individual datasets can be seen in Figures~\ref{fig:f1-bars} and~\ref{fig:auroc-bars}}.

\renewcommand{\arraystretch}{1.2}
\begin{tabular}{|l|l|c|c|c|c|}
\hline
\multicolumn{2}{|c|}{\multirow{3}{*}{\textbf{Methods}}}                          & \multicolumn{4}{c|}{Experimental Results} \\ \cline{3-6} 
\multicolumn{2}{|l|}{}      & \multicolumn{2}{c|}{Independent Test Sets} & \multicolumn{2}{c|}{Local Test Sets}  \\ \cline{3-6}
\multicolumn{2}{|l|}{}      & Mean F1 ($\pm$ SD)     & Mean AUC ($\pm$ SD)     & Mean F1 ($\pm$ SD)     & Mean AUC ($\pm$ SD)     \\ \hline
\multicolumn{2}{|l|}{\textbf{Centralized Model}}   & 0.9391 ($\pm$ 0.0210) & 0.9815 ($\pm$ 0.0082) & 0.9153 ($\pm$ 0.0555) & 0.9904 ($\pm$ 0.0051) \\ \hline
\multirow{3}{*}{\textbf{Federated Models}} 
& FedAvg~\cite{mcmahan2017communication}  & 0.8956 ($\pm$ 0.0455) & 0.9447 ($\pm$ 0.0186) & 0.8192 ($\pm$ 0.1329) & 0.9603 ($\pm$ 0.0314) \\ \cline{2-6}
& FedProx~\cite{li2018federated}    & 0.8579 ($\pm$ 0.0916) & 0.9234 ($\pm$ 0.0417) & 0.7800 ($\pm$ 0.1403) & 0.9360 ($\pm$0.0397) \\ \cline{2-6}
& FedDrAvg                          & \textbf{0.9102} ($\pm$ 0.0324) & \textbf{0.9542} ($\pm$ 0.0177) & \textbf{0.8482} ($\pm$ 0.0998) & \textbf{0.9645} ($\pm$ 0.0336) \\ \hline
\end{tabular}
\end{center}
\label{table:perf1}
\end{table*}
\begin{table*}[ht]
\centering
\begin{center}
\caption{3-fold experiments: average performance of the centralized and federated approaches on the held-out test sets of the centers participated in training and datasets of the Independent Centers (not participated in training). Performance given as the mean and standard deviation of the F1 scores and AUROC values.}

\renewcommand{\arraystretch}{1.2}
\begin{tabular}{|l|l|c|c|c|c|}
\hline
\multicolumn{2}{|c|}{\multirow{3}{*}{\textbf{Methods}}}                        & \multicolumn{4}{c|}{3-fold Experimental Results}  \\ \cline{3-6} 
\multicolumn{2}{|l|}{}          & \multicolumn{2}{c|}{Independent Test Sets} & \multicolumn{2}{c|}{Local Test Sets}         \\ \cline{3-6} 
\multicolumn{2}{|l|}{}          & Mean F1 ($\pm$ SD)     & Mean AUC ($\pm$ SD)     & Mean F1  ($\pm$ SD)     & Mean AUC ($\pm$ SD)     \\ \hline
\multicolumn{2}{|l|}{\textbf{Centralized Model}}    & 0.9088 ($\pm$ 0.0969) & 0.9688 ($\pm$ 0.0535) & 0.9274 ($\pm$ 0.0439) & 0.9883 ($\pm$ 0.0085) \\ \hline
\multirow{3}{*}{\textbf{Federated Models}} 
& FedAvg~\cite{mcmahan2017communication}  & 0.8289 ($\pm$ 0.1467) & 0.9166 ($\pm$ 0.0856) & 0.8478 ($\pm$ 0.1061) & 0.9514 ($\pm$ 0.0307) \\ \cline{2-6} 
& FedProx~\cite{li2018federated}          & 0.8341 ($\pm$ 0.1477) & 0.9145 ($\pm$ 0.0895) & 0.8591 ($\pm$ 0.0940) & 0.9558 ($\pm$ 0.0310) \\ \cline{2-6} 
& FedDrAvg      & \textbf{0.8428} ($\pm$ 0.1458) & \textbf{0.9231} ($\pm$ 0.0916) & \textbf{0.8692}  ($\pm$ 0.0838) & \textbf{0.9597} ($\pm$ 0.0239) \\ \hline
\end{tabular}
\end{center}
\label{table:perf2}
\end{table*}


\subsubsection{Limitations of local models}
\label{ssec:limitationsoflocal}

The F1 score of each locally trained model (rows) on each local held-out test set and on independent test set given as heatmaps in Fig.~\ref{fig:f1s-localmodels} and Fig.~\ref{fig:f1s-localmodels-indep}. Comparing federated models (Fig.~\ref{fig:f1-bars}) with locally trained models, we observe that, on the individual local test sets, the federated methods, especially our proposed method, perform better than most of the locally trained models on the held-out test set of that center. This strengthens the motivation to put federated learning into practise. Comparing locally trained models with each other, we see that some of the locally trained models are indeed do best on the test set of the same client they are trained on (CM, A6, DY, G4, AA). Surprisingly we observe this is not the case for majority of them. For example, we see that the model locally trained on the training data of DC give the best F1 results compared to other local models on the test sets of AY and AH. Likewise, among other local models, the locally trained model on A6 is best for AF, CK and DC; and locally trained model on AA is best for AG. These results point to complexity of the underlying relationships between the different local models. They also supports our initial motivation about not being able to measure the individual contributions of each local dataset without sharing the datasets.

\subsubsection{Experimental analysis}
\label{ssec:experimental}

To compare different approaches, F1 scores and AUROC values are calculated for each comparison model on 21 different test sets (held-out data of training centers and data of independent centers). 

In table~\ref{table:perf1}, average performance of the centralized and federated approaches are shown. In the table, performance is given as the mean and the standard deviation of the F1 metric on the local and independent test sets data. As expected, the {\em Centralized} model, which is trained on all of the training data in a classical way, gives the best mean F1 score among all of the comparison models. Compared to other federated approaches, proposed \textit{FedDropoutAvg} model (with $cdr=0.2$ and $fdr=0.3$) produced the most competitive result with the {\em Centralized} model. 

In Fig.~\ref{fig:f1-bars} and Fig.~\ref{fig:auroc-bars}, we see the performance comparison of federated approaches on the individual local test sets and for individual independent sets. In these results, we see that proposed \textit{FedDropoutAvg} method consistently better than other federated approaches on these individual testing sets. This success of our proposed approach is attributed to the proposed client dropout and clients' model weights drop-out mechanism. Our approach indeed helps avoiding over-fitting to individual local datasets.

In the held-out test sets of center AY and AA, the ratio of tumor vs non-tumor tiles is very imbalanced (Table~\ref{table:dataset}). This could be the reason why the F1 performances of all of the federated approaches on the test sets of these centers are lower compared to other centers (Fig.~\ref{fig:f1-bars}). Although our approach still has better performance than the other federated models.

\subsubsection{3-fold experimental analysis}
\label{ssec:3fold}
We also performed 3-fold cross validation experiments. We divided the 21 centers into three splits, where each split contains 7 centers. At each fold, the centers in the two splits have participated in training while the centers in the remaining one split are used as the independent test sets. Similar to the previous experiments, at each fold, we have calculated the performance metrics from the independent centers and from the held-out test sets of the centers participated in training. The results of these experiments can be seen in table~\ref{table:perf2}. These results further confirm that our proposed approach is superior than other federated approaches in terms of F1 and AUC values on both local and independent test sets.

\subsubsection{Experimental analysis for training with different {\em cdr} and {\em fdr} parameters}

To understand the effects of $cdr$ and $fdr$ parameters, we also trained the proposed \textit{FedDropoutAvg} method with different parameters. Keeping in mind that the model federatively trained with $cdr=0$ and $fdr=0$ corresponds to the FedAvg method in the literature, we see that selecting greater than zero values for both of the parameters provides gains in F1 score. 

\begin{figure}
\centering
\begin{tabular}{@{~}c@{~}}
\includegraphics[width=0.70\linewidth]{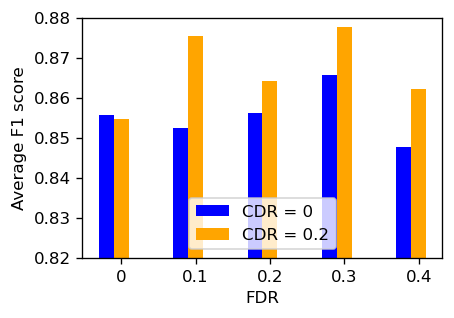} \\
(a) \\
\includegraphics[width=0.70\linewidth]{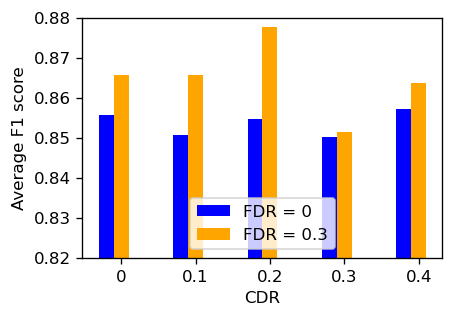} \\
(b) \\
\end{tabular}

\caption{Performance results of proposed method trained with different $cdr$ or $fdr$ parameters. Performance given as the average of the F1 scores on all of the test sets for the proposed \textit{FedDropoutAvg} method. For demonstration purposes here, performance is obtained on the held-out test sets of the centers participated in training and datasets of the Independent Centers (not participated in training). The best parameters are originally selected on the validation sets of the training centers as $cdr=0.2$ and $fdr=0.3$. Note that \textit{FedDropoutAvg} with $cdr=0$ and $fdr=0$ is same with the \textit{FedAvg} method from the literature.}
\label{fig:params}
\end{figure}

In Fig.~\ref{fig:params}(a) performance results of models trained with different $fdr$ parameters when $cdr$ parameter is equal to 0 (as in \textit{FedAvg}) and equal to 0.2 (as the reported model here) can be seen. Here, we see that the models trained with $cdr=0.2$ give close or better results than their counterparts ($cdr=0$).

Likewise, in Fig.~\ref{fig:params}(b), performance results of models trained with different $cdr$ parameters when $fdr$ parameter is equal to 0 (as in \textit{FedAvg}) and equal to 0.3 (as the reported model here) are presented. Here, we also see that the models trained with $fdr=0.3$ give better results than their counterparts ($fdr=0$).

Additionally, in Fig.~\ref{fig:params}(b), we can argue that in standard federated learning (i.e., when we do not use federated dropout, $fdr=0$), if we decrease the number of clients participating at each round (i.e., increasing the client dropout rate, $cdr$) the performance does not differ noticeably.

\subsubsection{Qualitative analysis on WSI level}
\label{ssec:qualitative}
In Fig.~\ref{fig:segmentation-examples}, qualitative results of different methods on WSI level are presented with respective slide-level F1 scores. Centralized, federated and local training methods are compared. The WSIs in this figure are from four different independent testing centers (AD, F4, NH, D5), thus the models are not trained on these centers' dataset. The best local models are selected based on comparison of all of the locally trained models by the slide-level F1 metrics. 

For all of the WSIs, our proposed method has given better results than other federated approaches (FedAvg, FedProx) both qualitatively and in terms of F1 score. In the last three columns, we see that best of the locally trained models give better results than federated approaches. The results in the last column can be regarded as a fail case for all of the federated models, although it should be noted that our model still gives better results.


\newcommand{\inc}[1]{\includegraphics[align=c,width=.16\textwidth]{#1}} 
\newcommand{\incc}[1]{\includegraphics[align=c,width=.24\textwidth]{#1}} 
\newlength{\himg} 
\setlength{\himg}{-31pt} 
\newlength{\htext} 
\setlength{\htext}{-23pt} 

\begin{figure*}[!htbp]
\centering
\scriptsize{
\setlength{\tabcolsep}{-4pt}
\begin{tabular}{cccccc}  
\\[-15pt]
& AD-5900-01Z-00-DX1 & F4-6570-01Z-00-DX1 & NH-A50U-01Z-00-DX1 & D5-6927-01Z-00-DX1\\[\htext]
WSI &
\inc{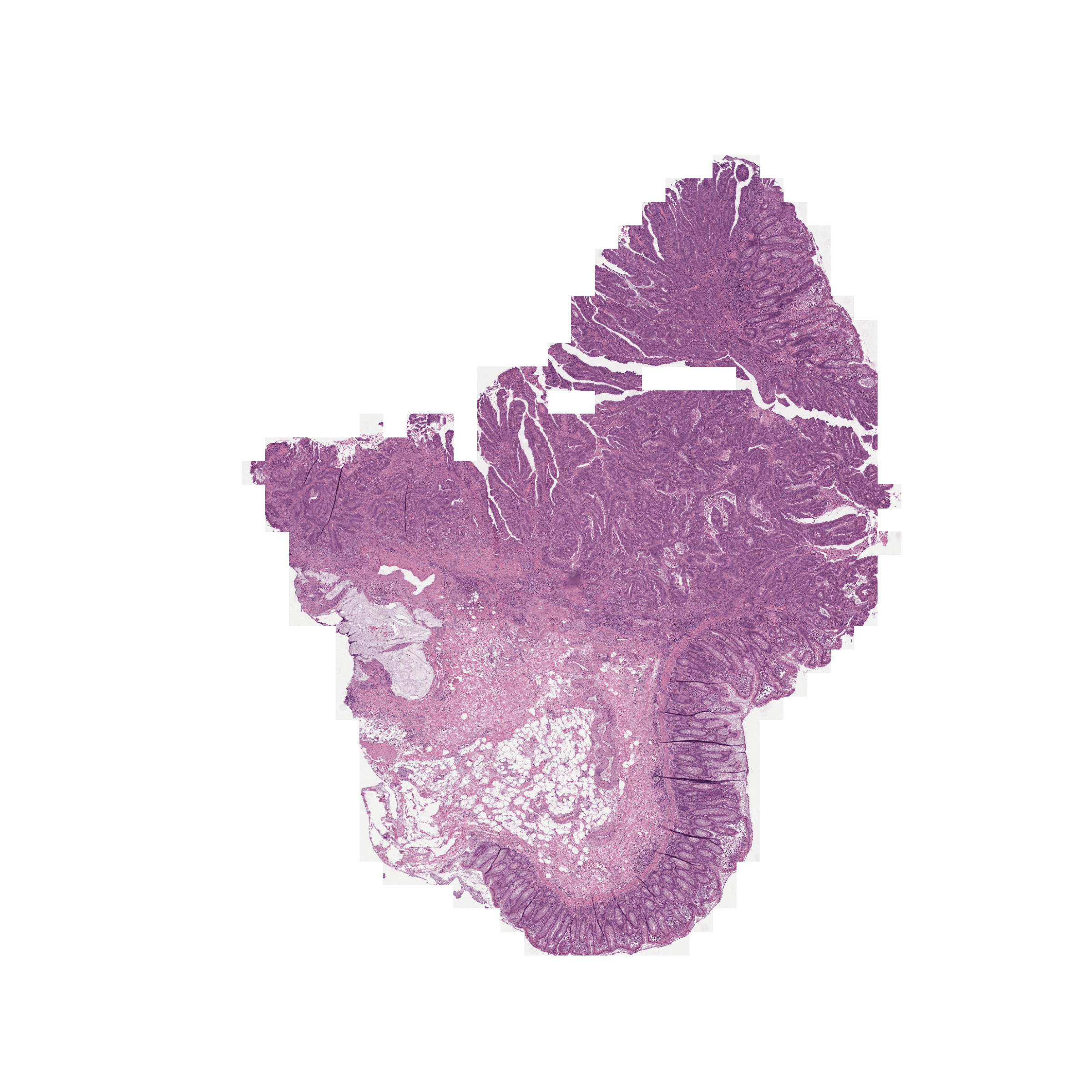} &
\inc{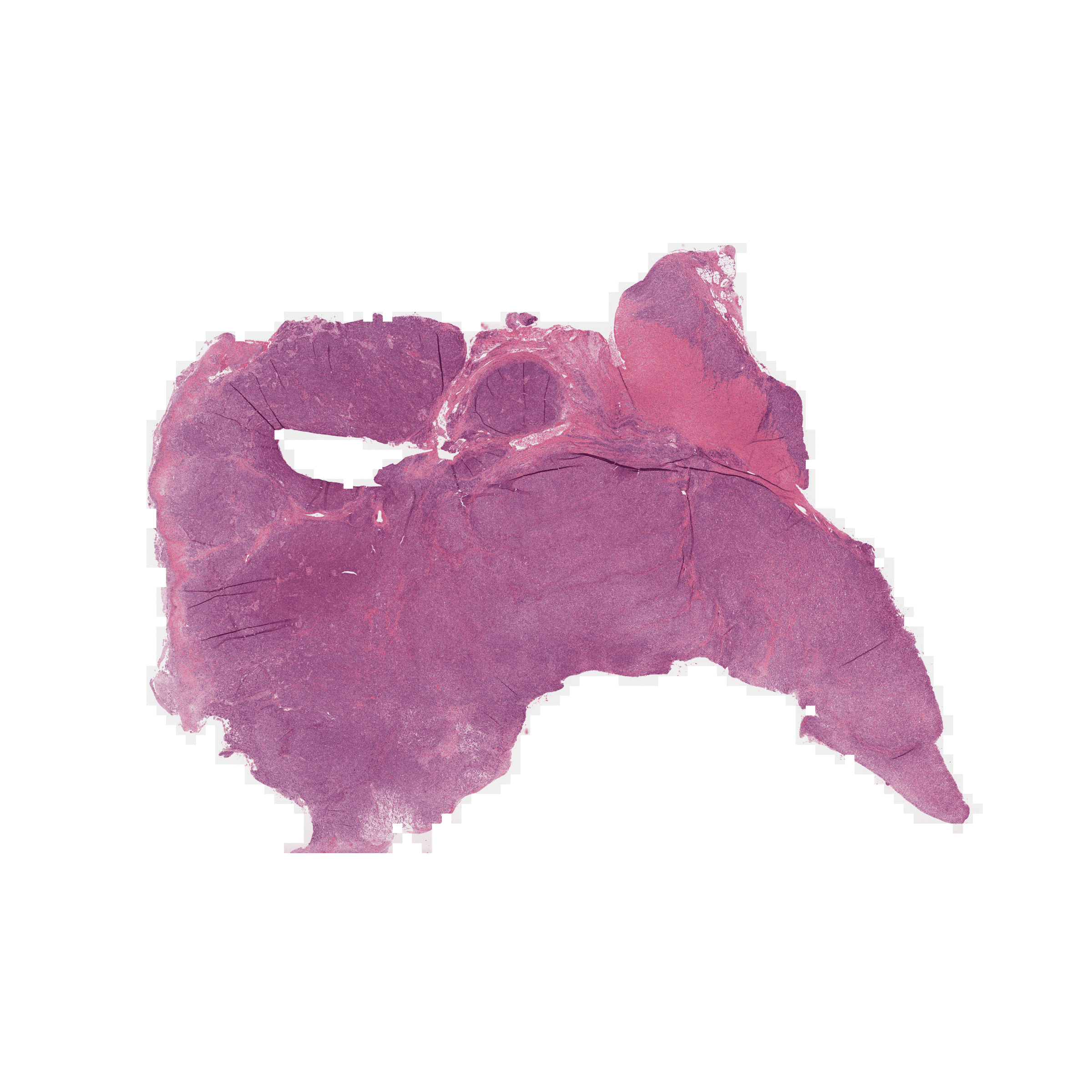} &
\inc{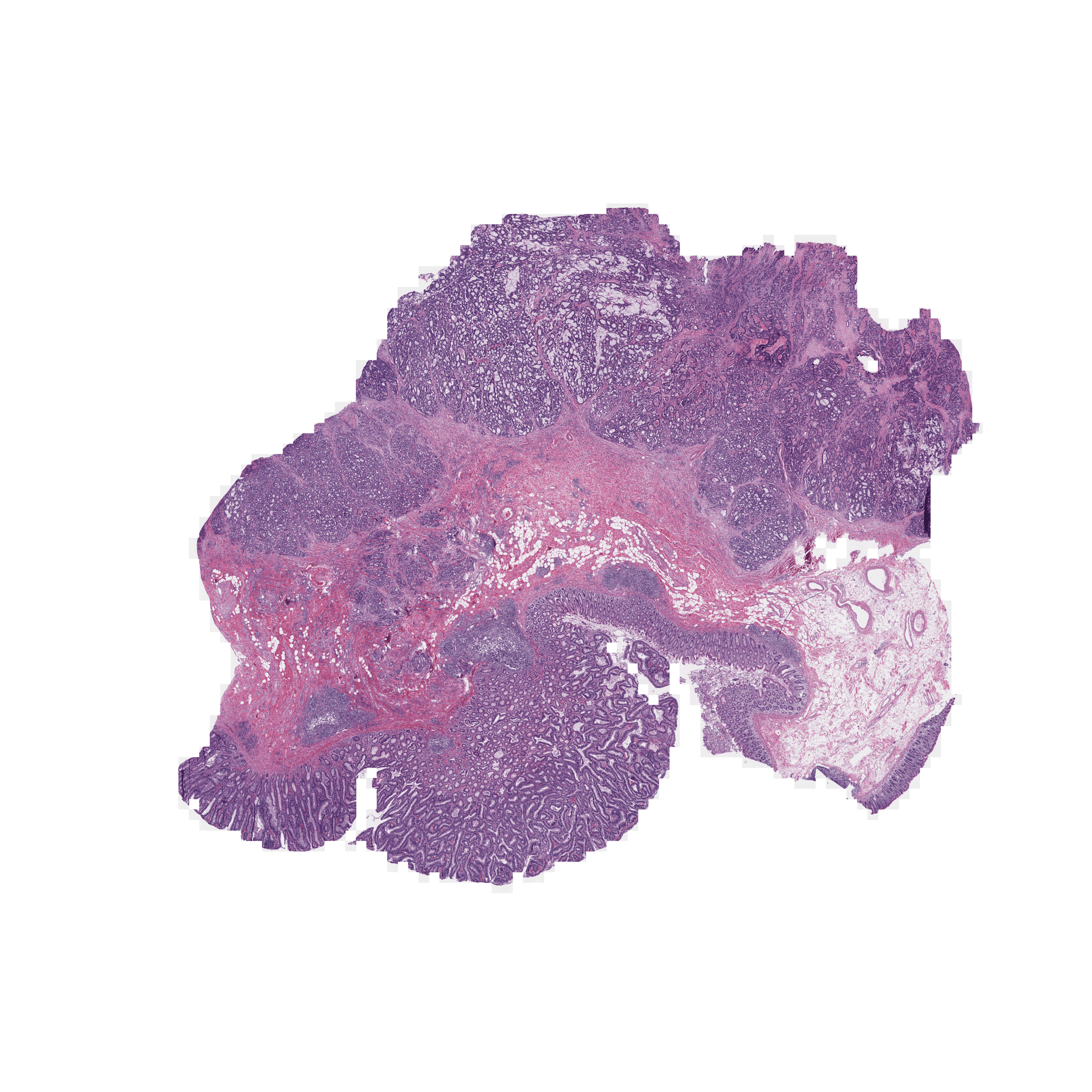} & \incc{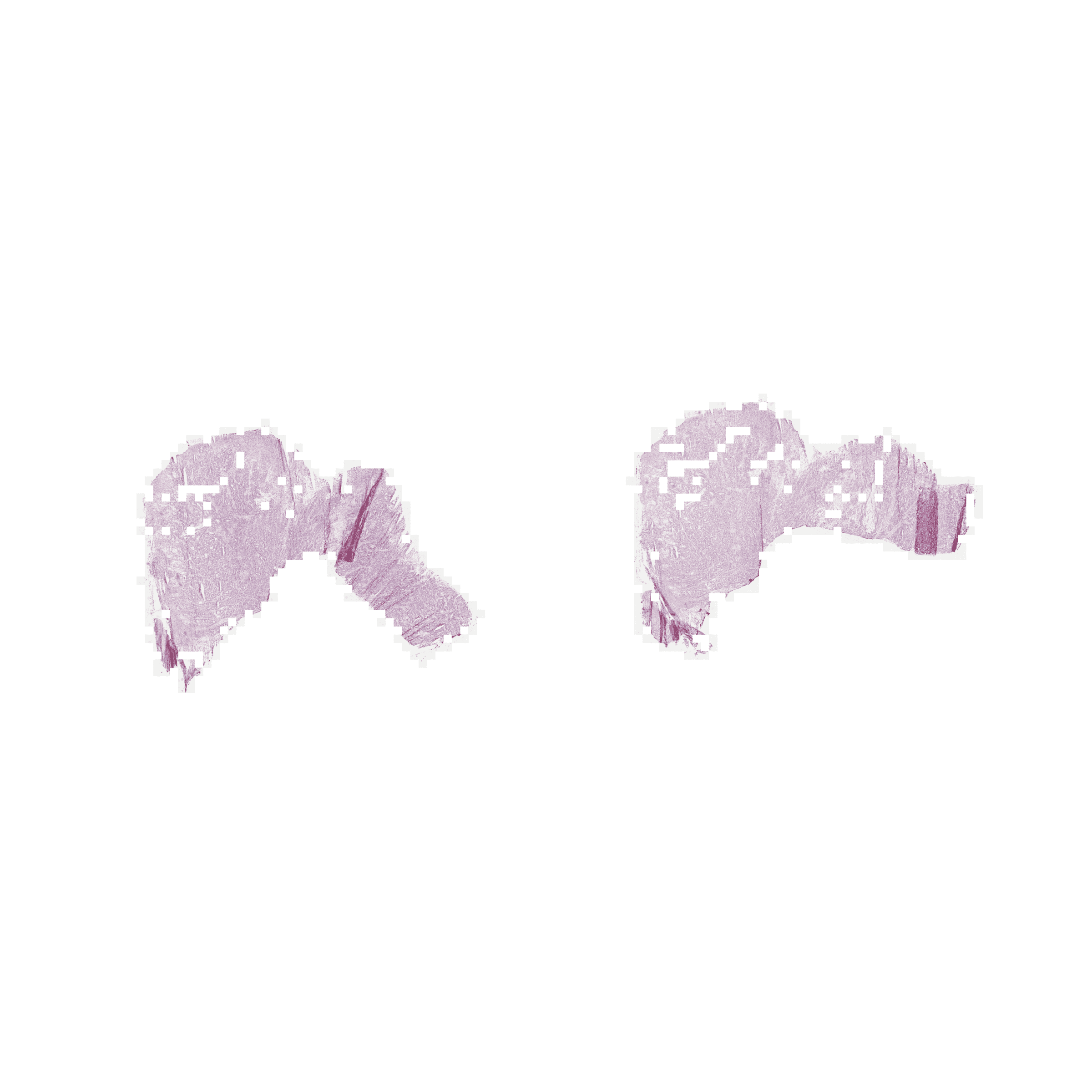} \\[-55pt]
GT &
\inc{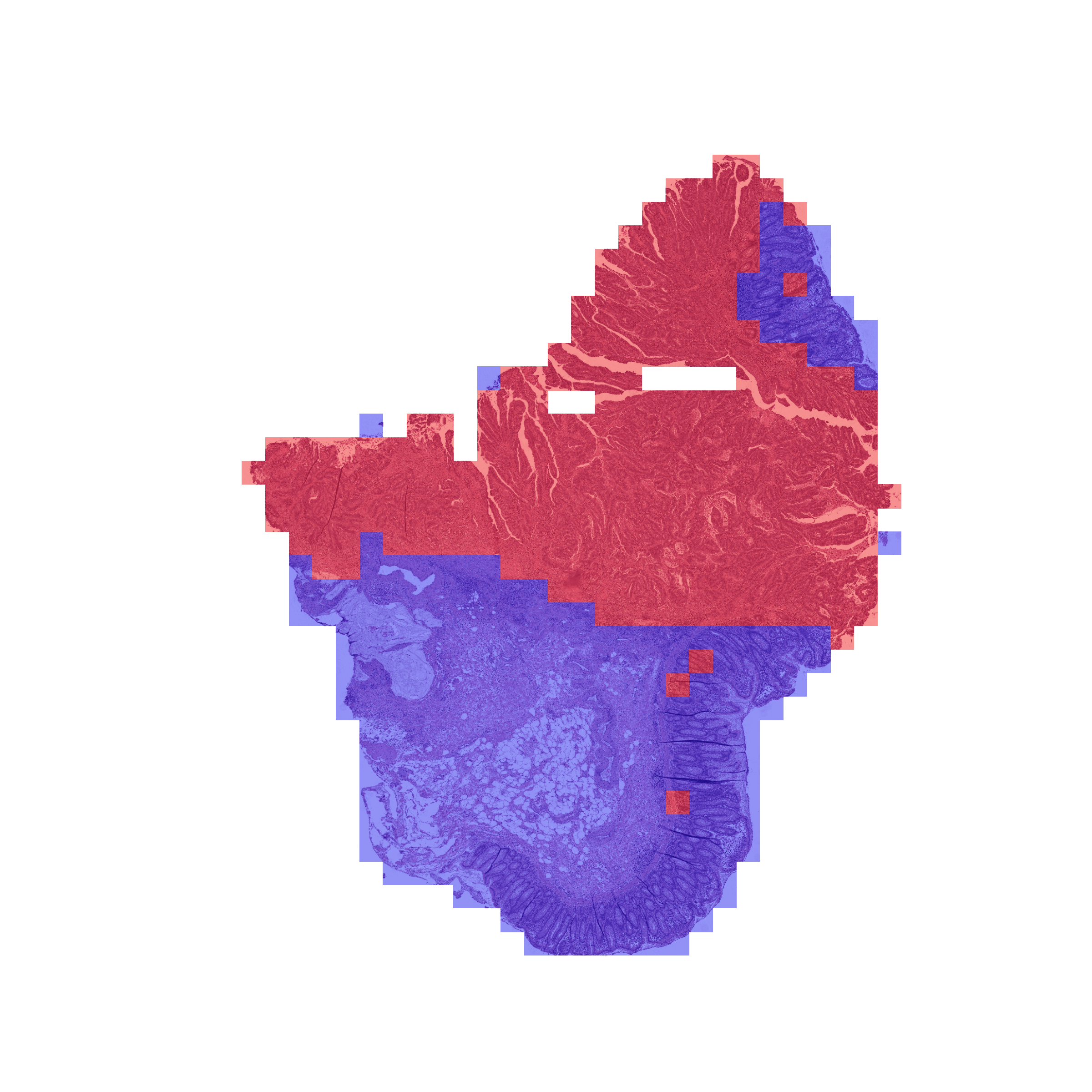} &
\inc{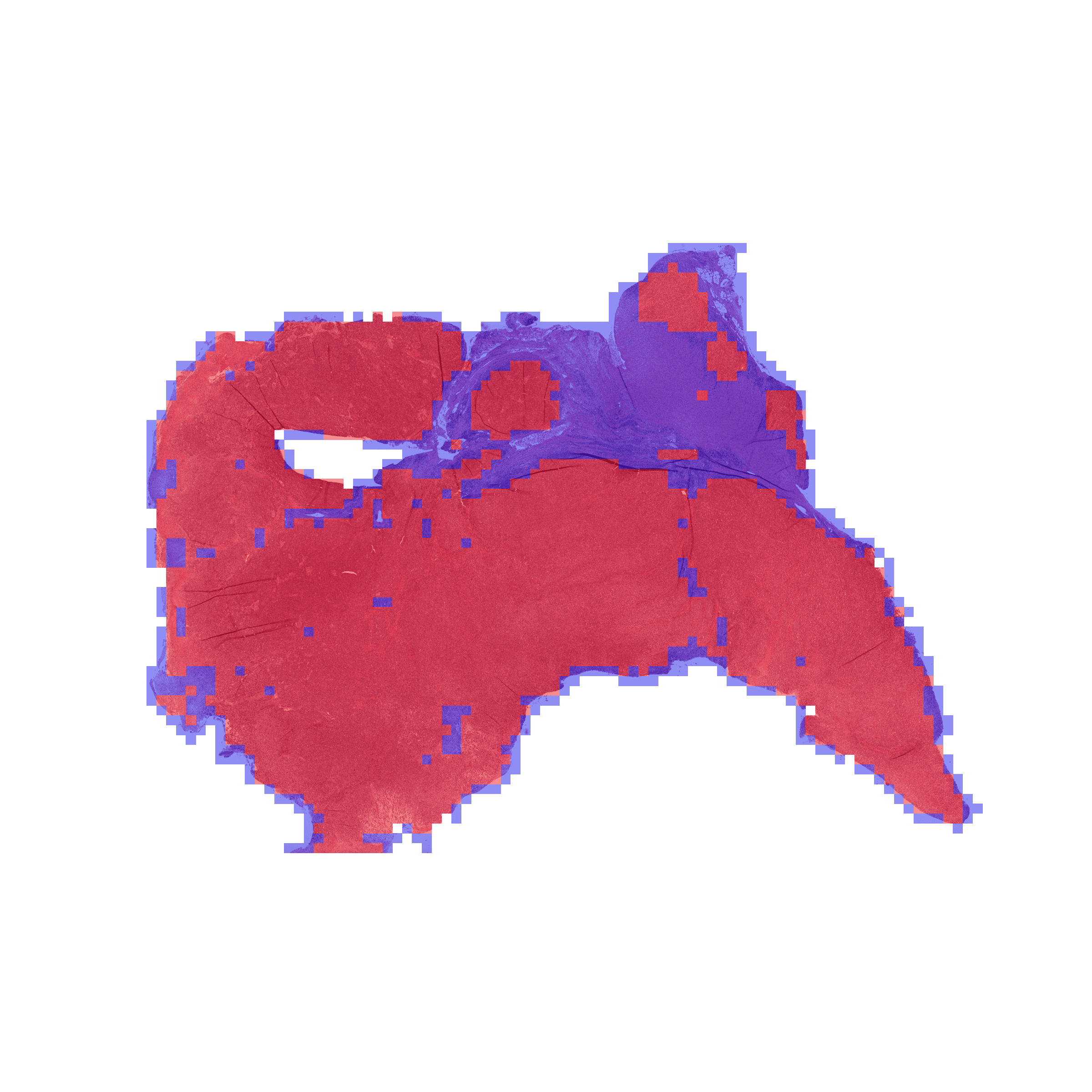} &
\inc{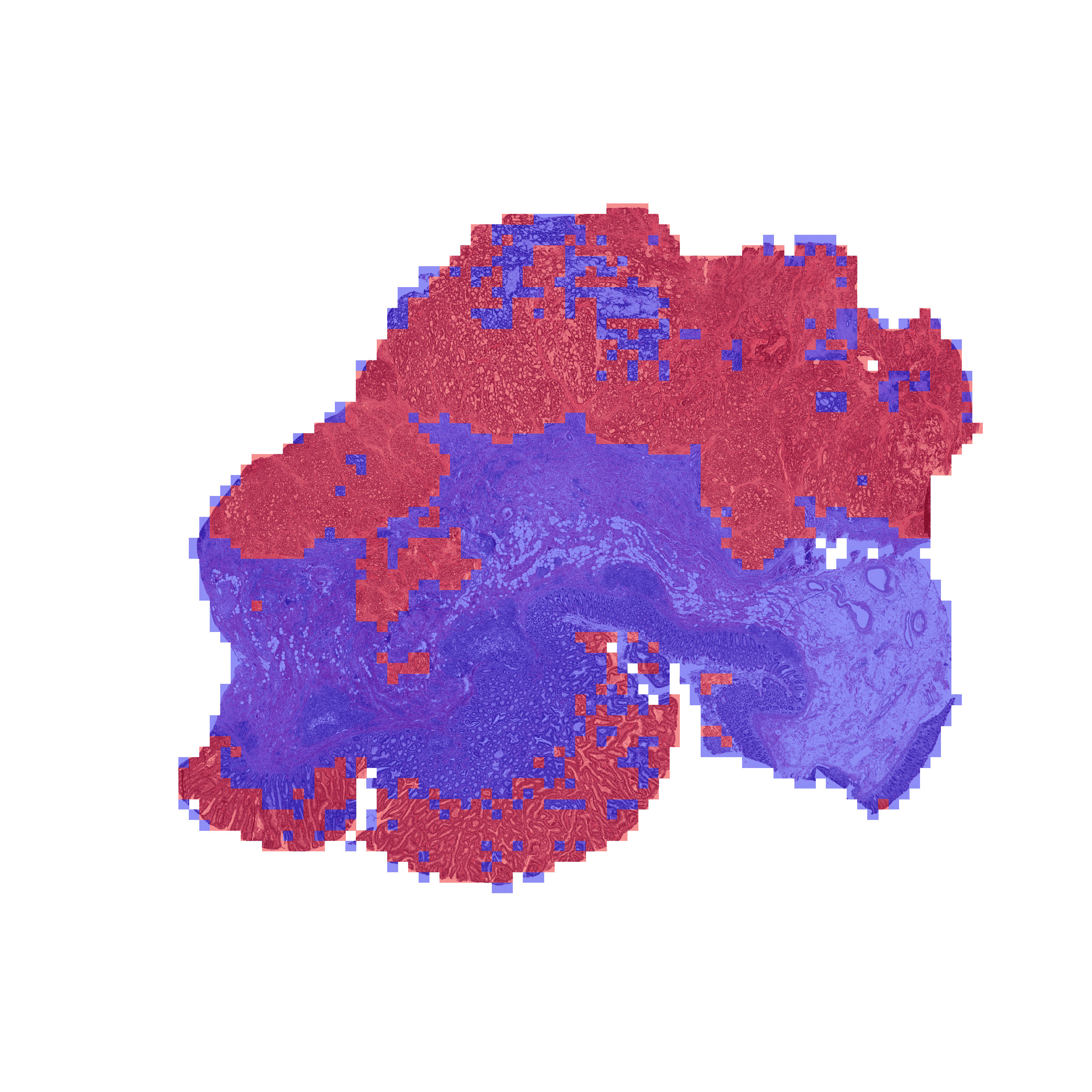} &
\incc{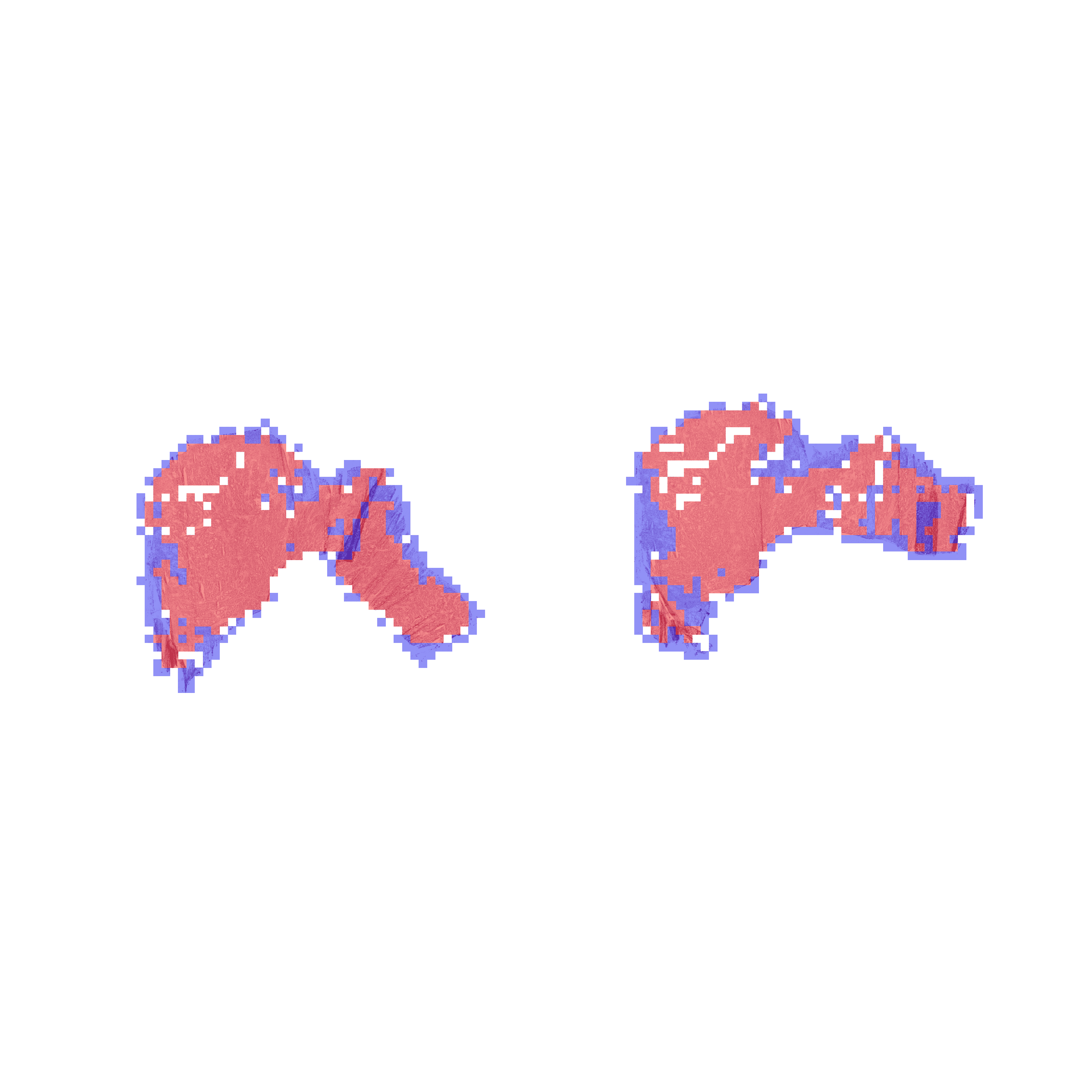} \\[-55pt]
Centralized &
\inc{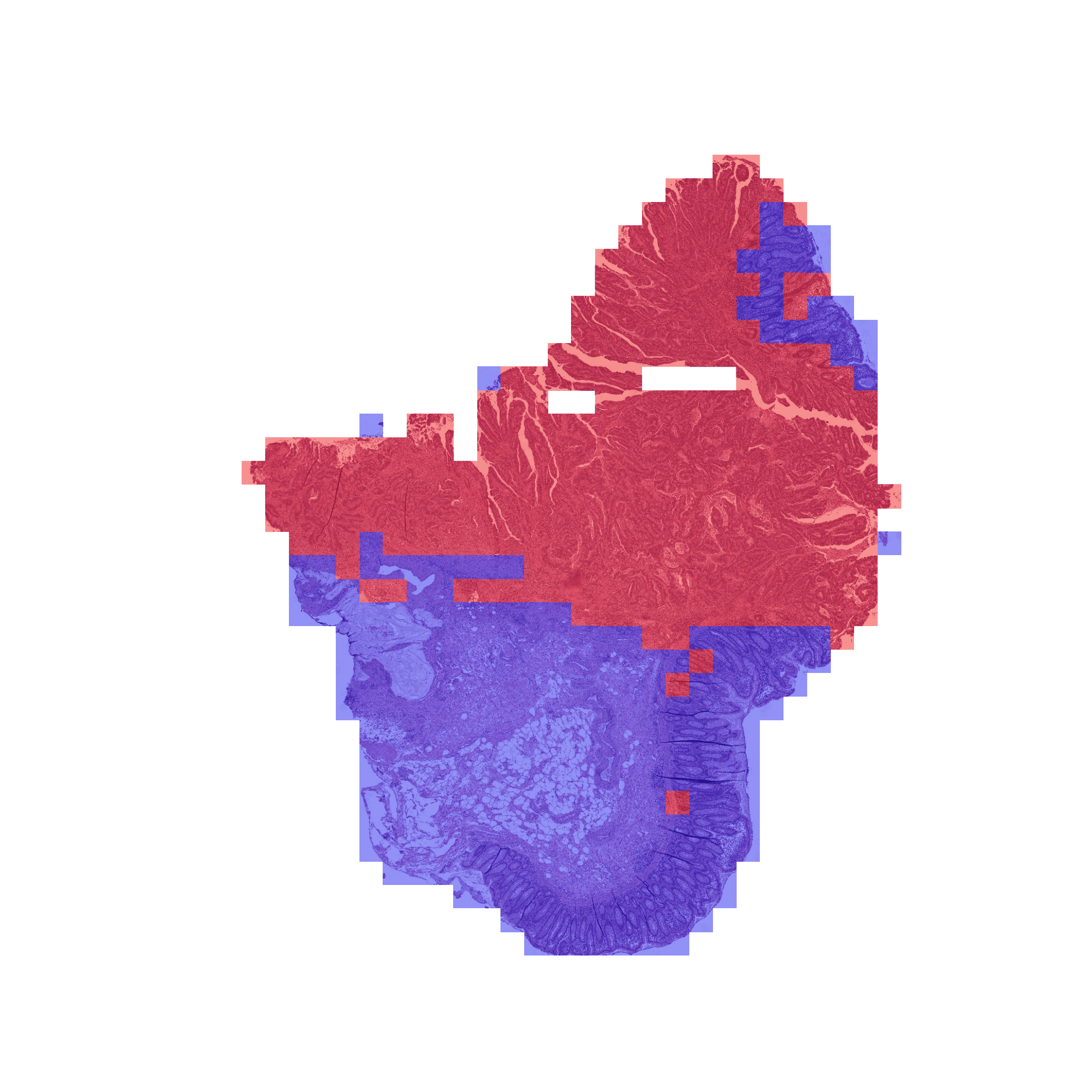} &
\inc{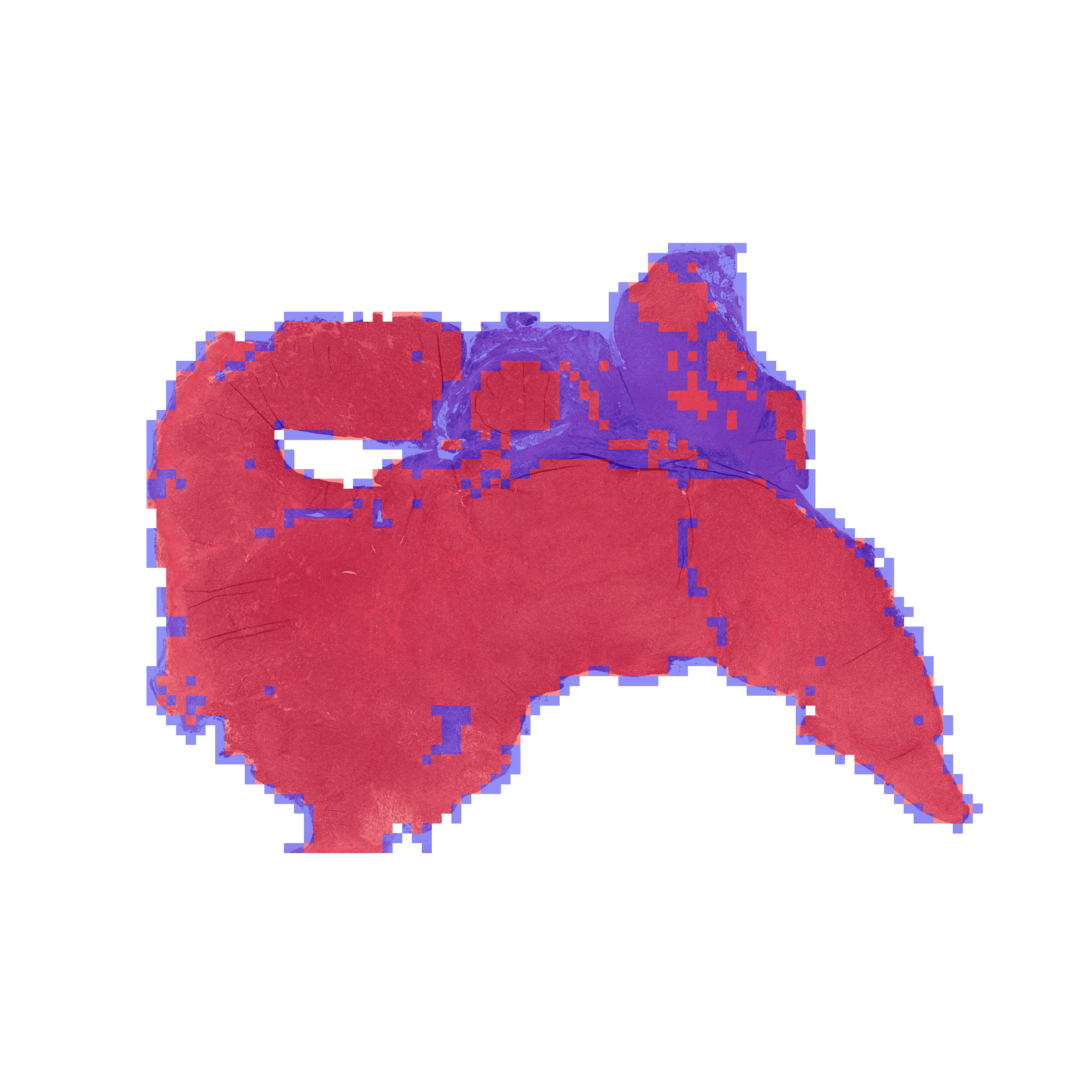} &
\inc{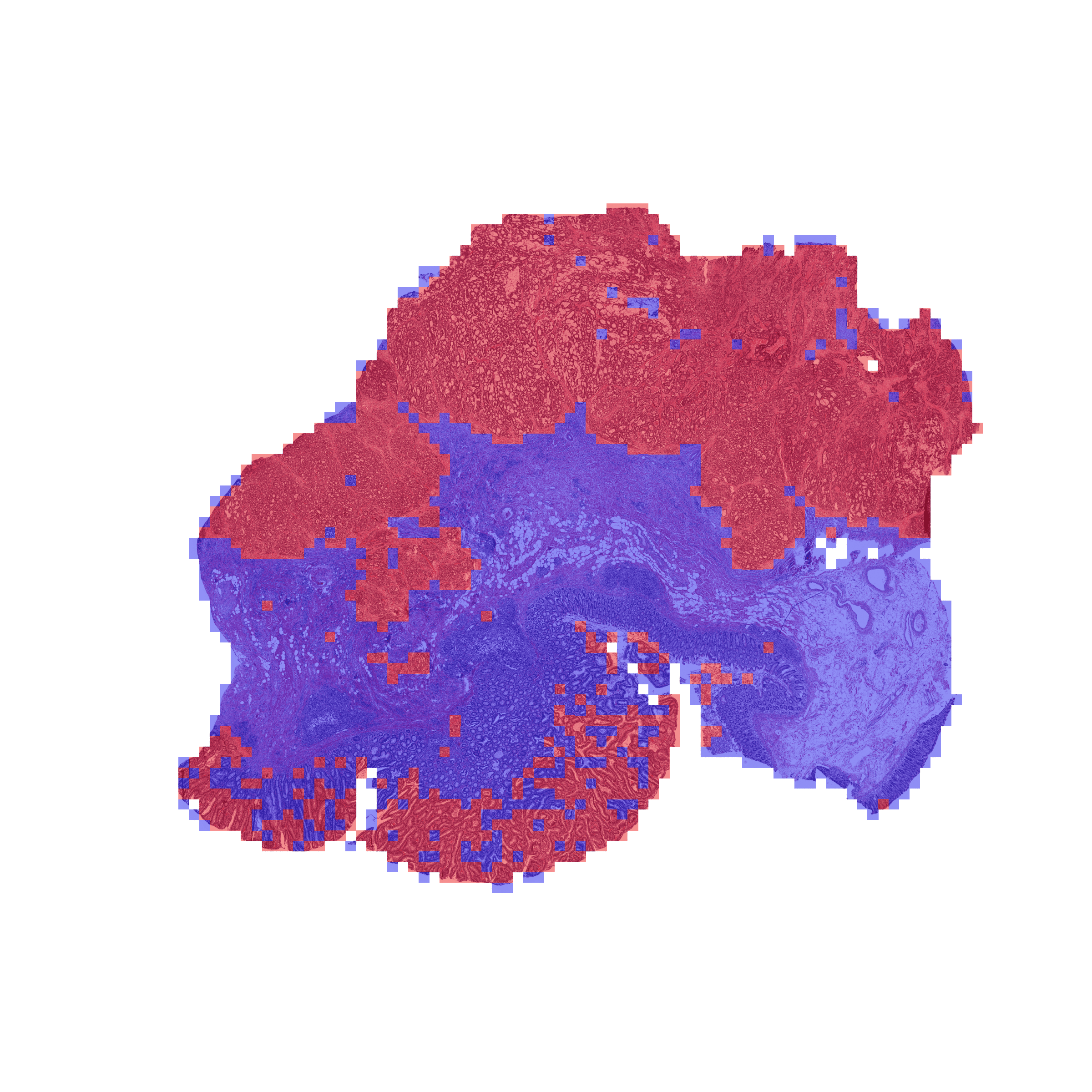} &
\incc{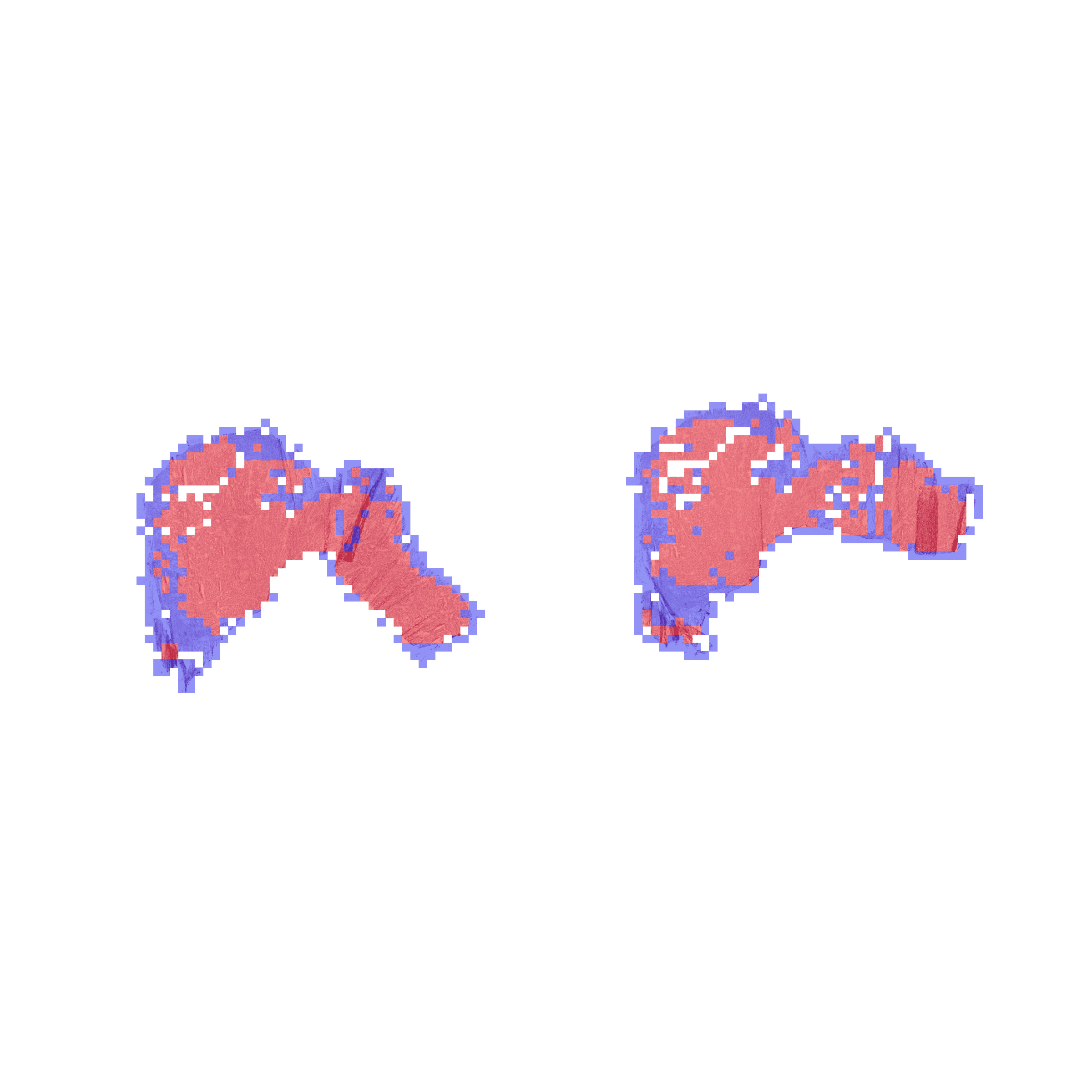} \\[\himg]
& F1=0.97 & F1=0.87 & F1=0.92 & F1=0.77 \\[\htext]
Ours &
\inc{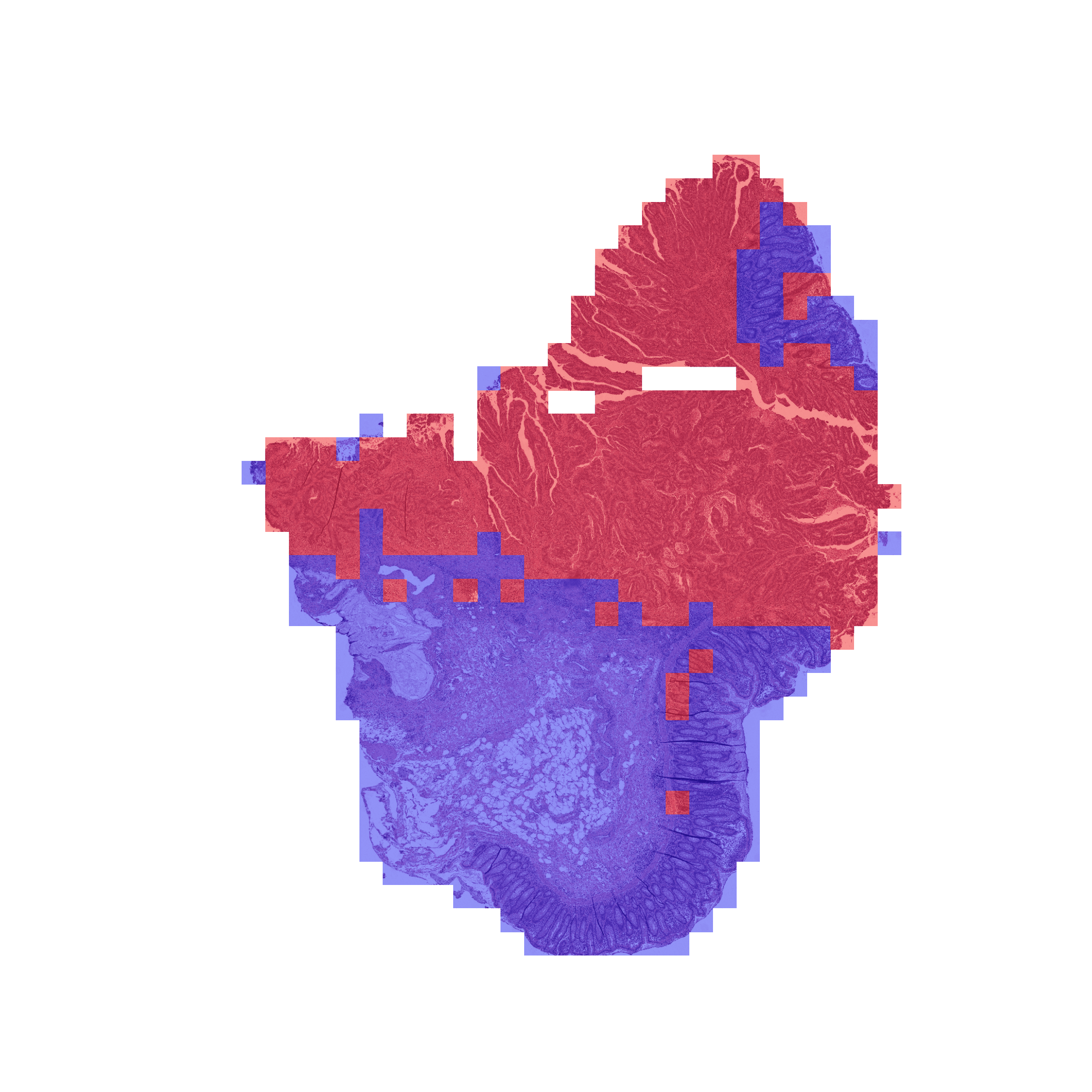} &
\inc{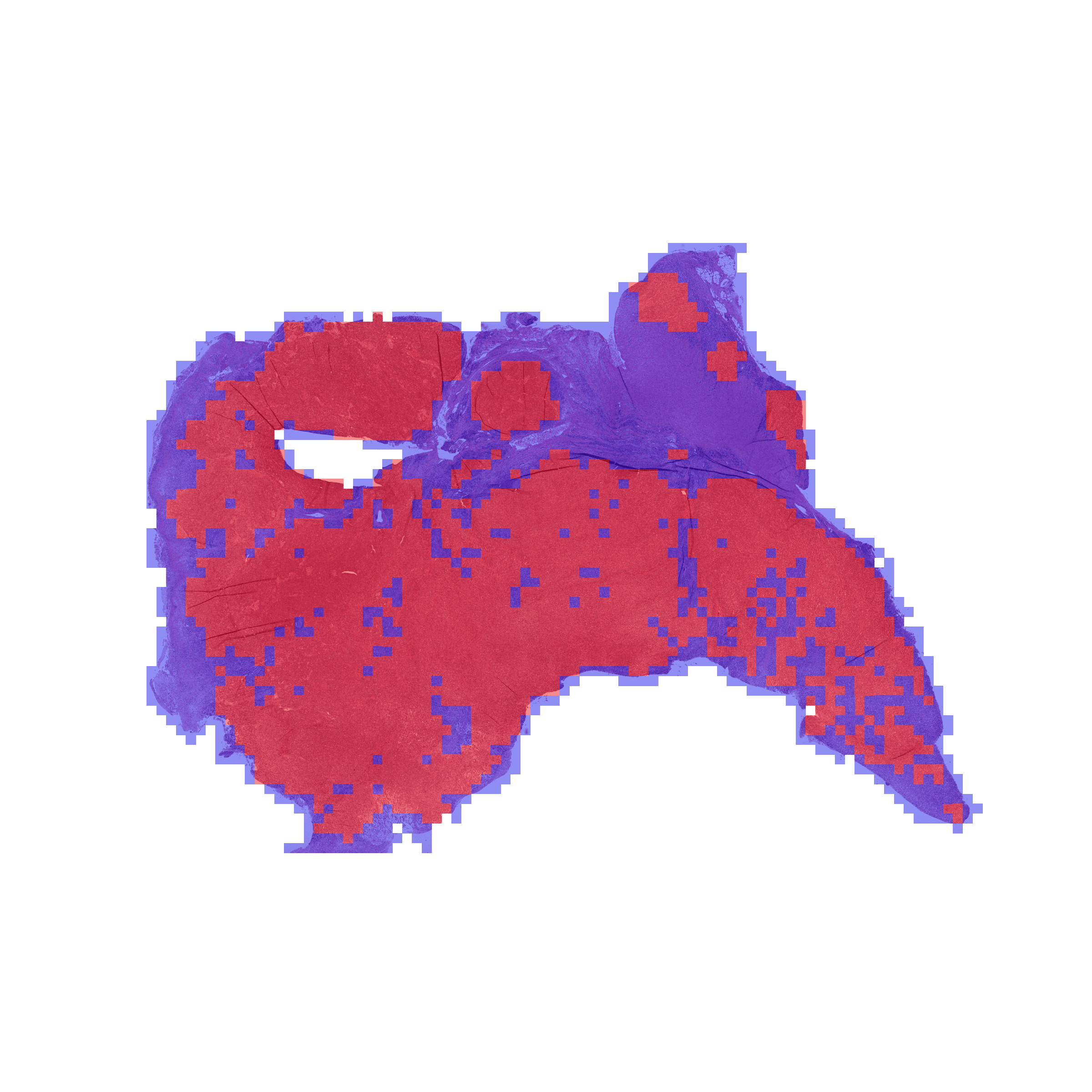} &
\inc{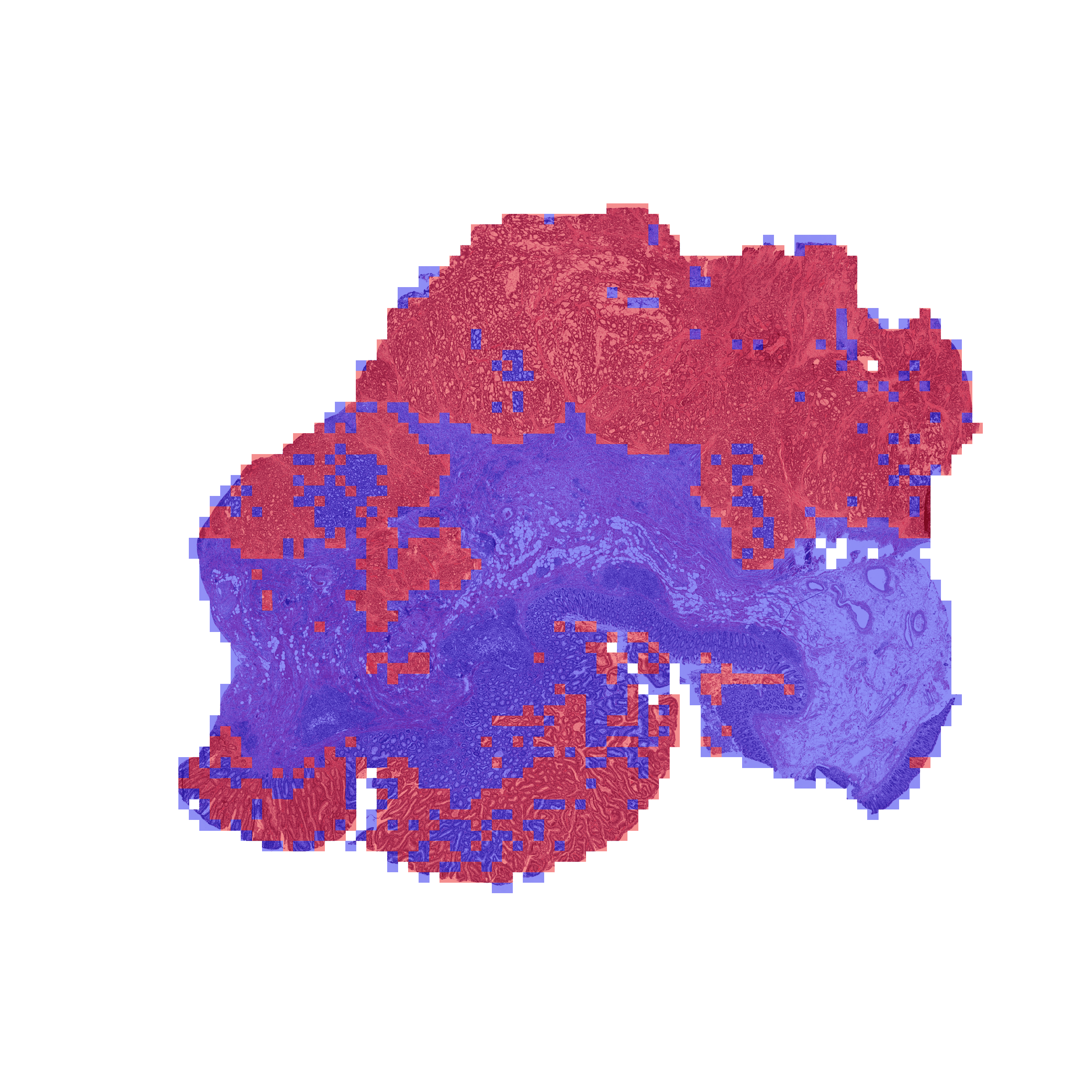} &
\incc{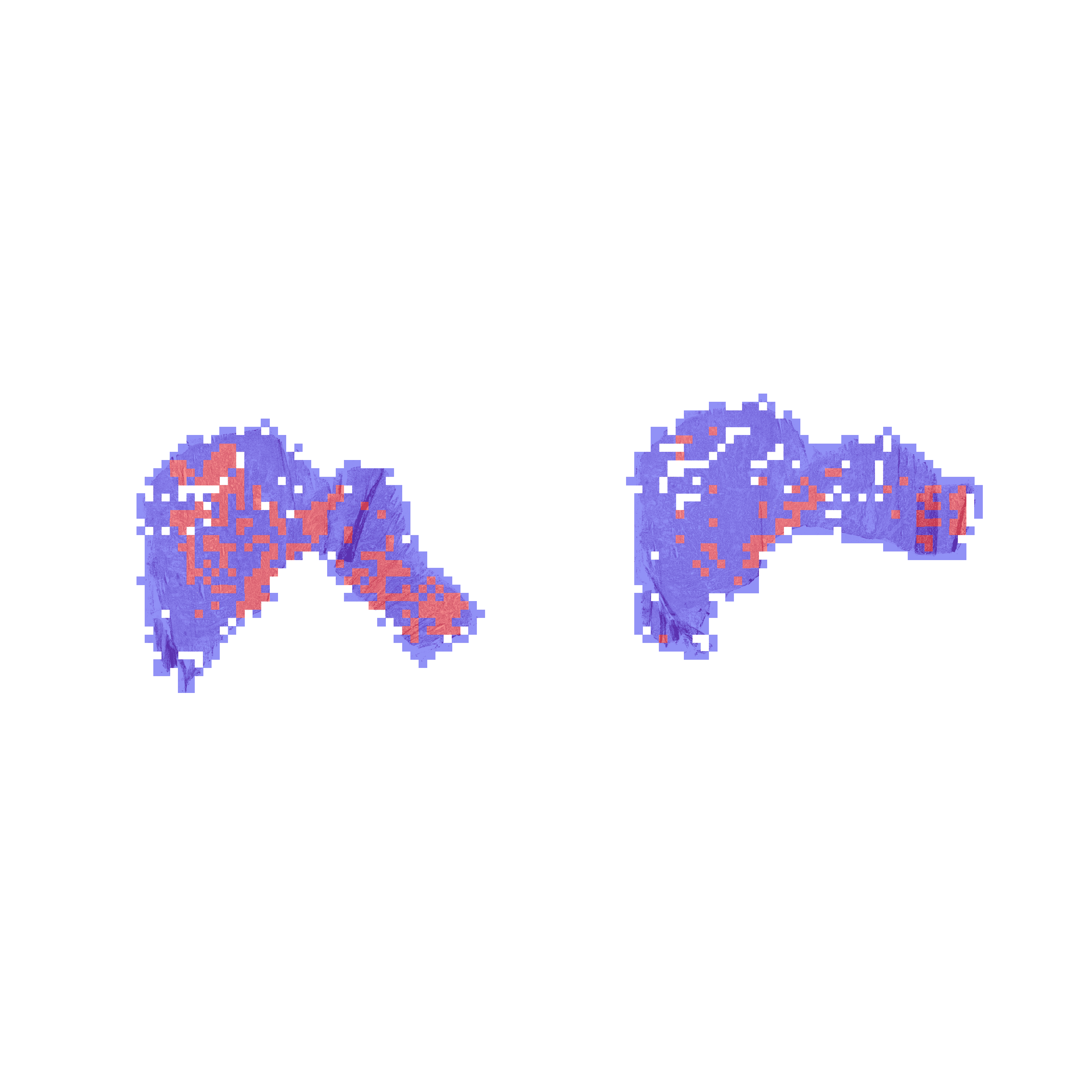} \\[\himg]
& F1=0.96 & F1=0.78 & F1=0.87 & F1=0.56 \\[\htext]
FedAvg &
\inc{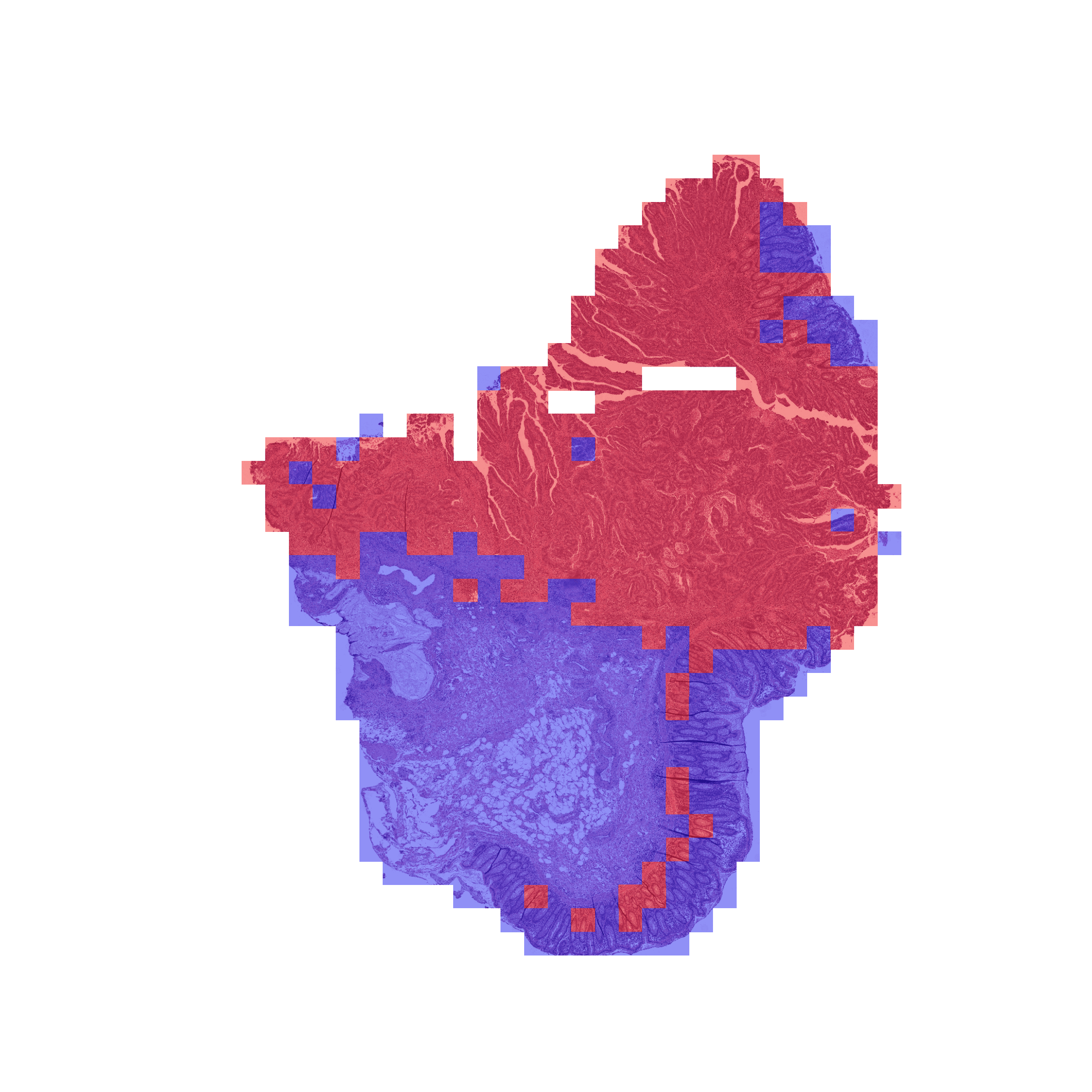} &
\inc{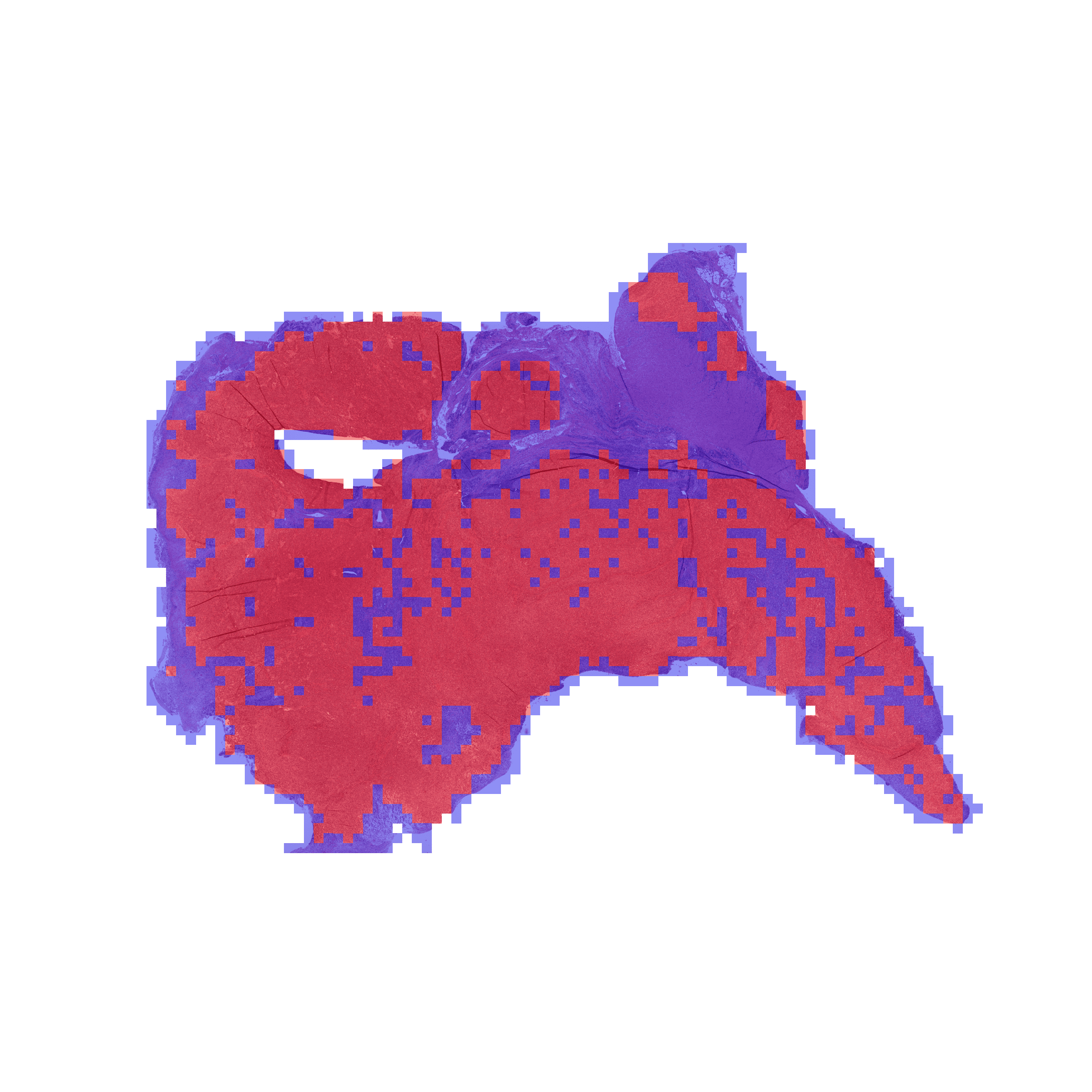} &
\inc{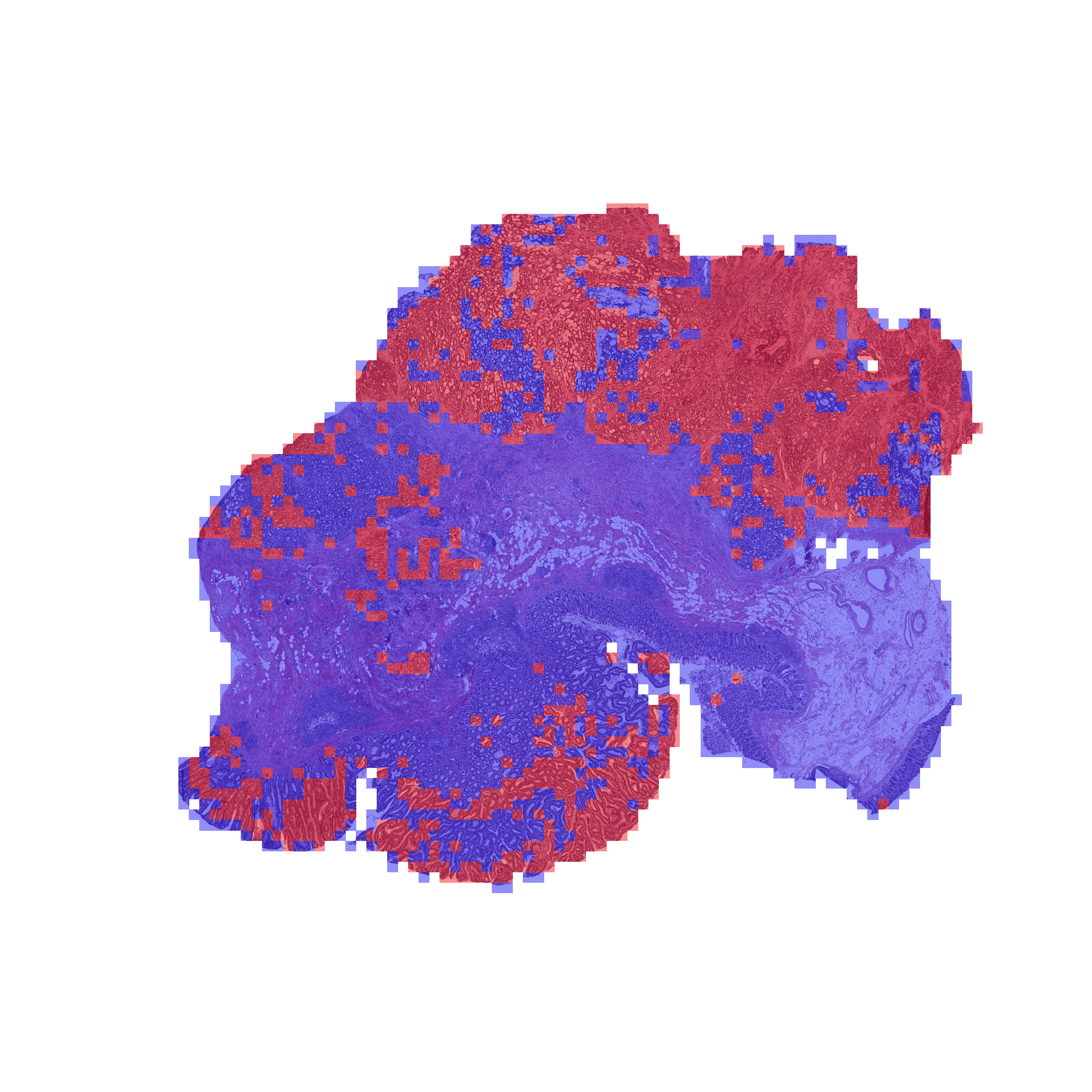} &
\incc{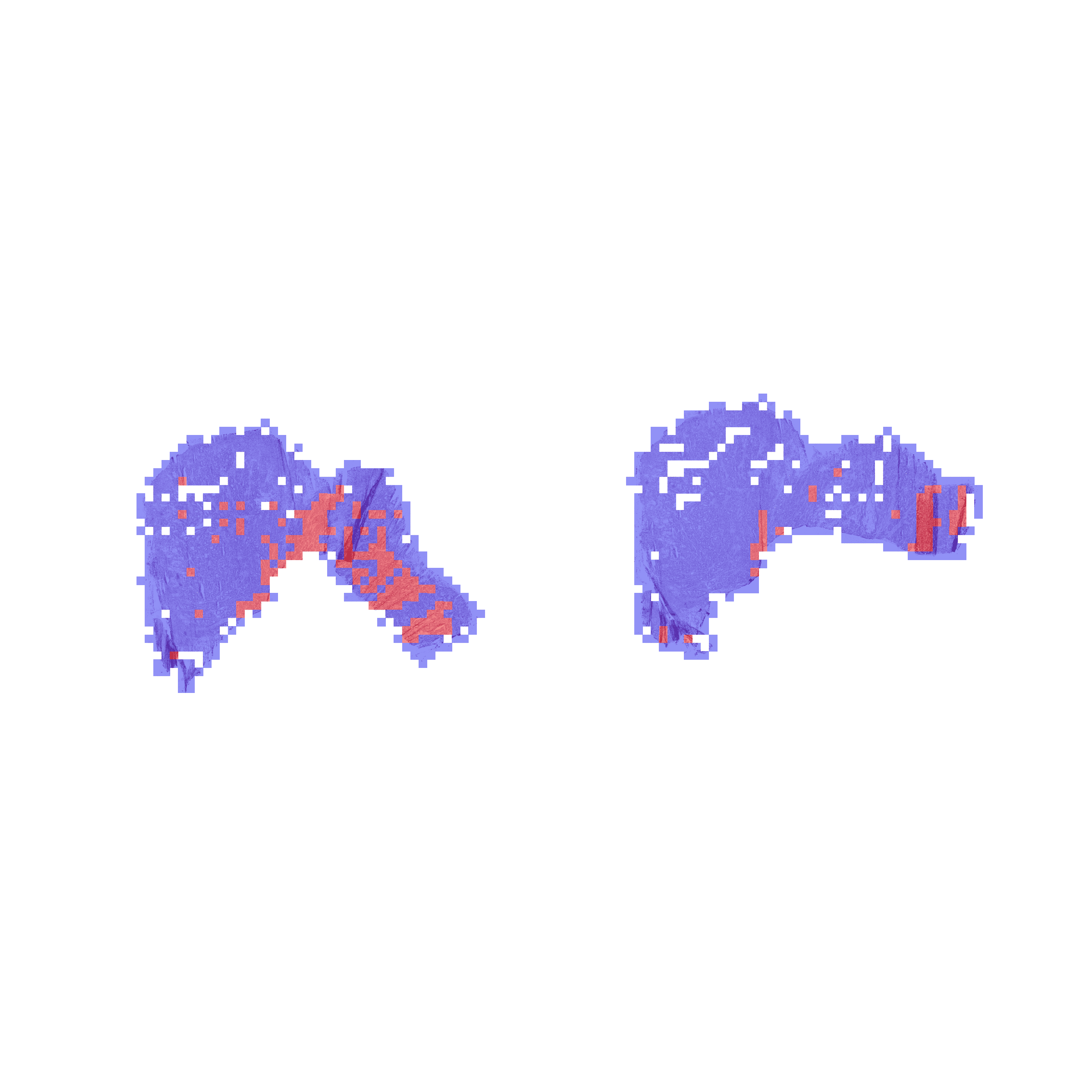} \\[\himg]
& F1=0.93 & F1=0.76 & F1=0.84 & F1=0.52 \\[\htext]
FedProx & 
\inc{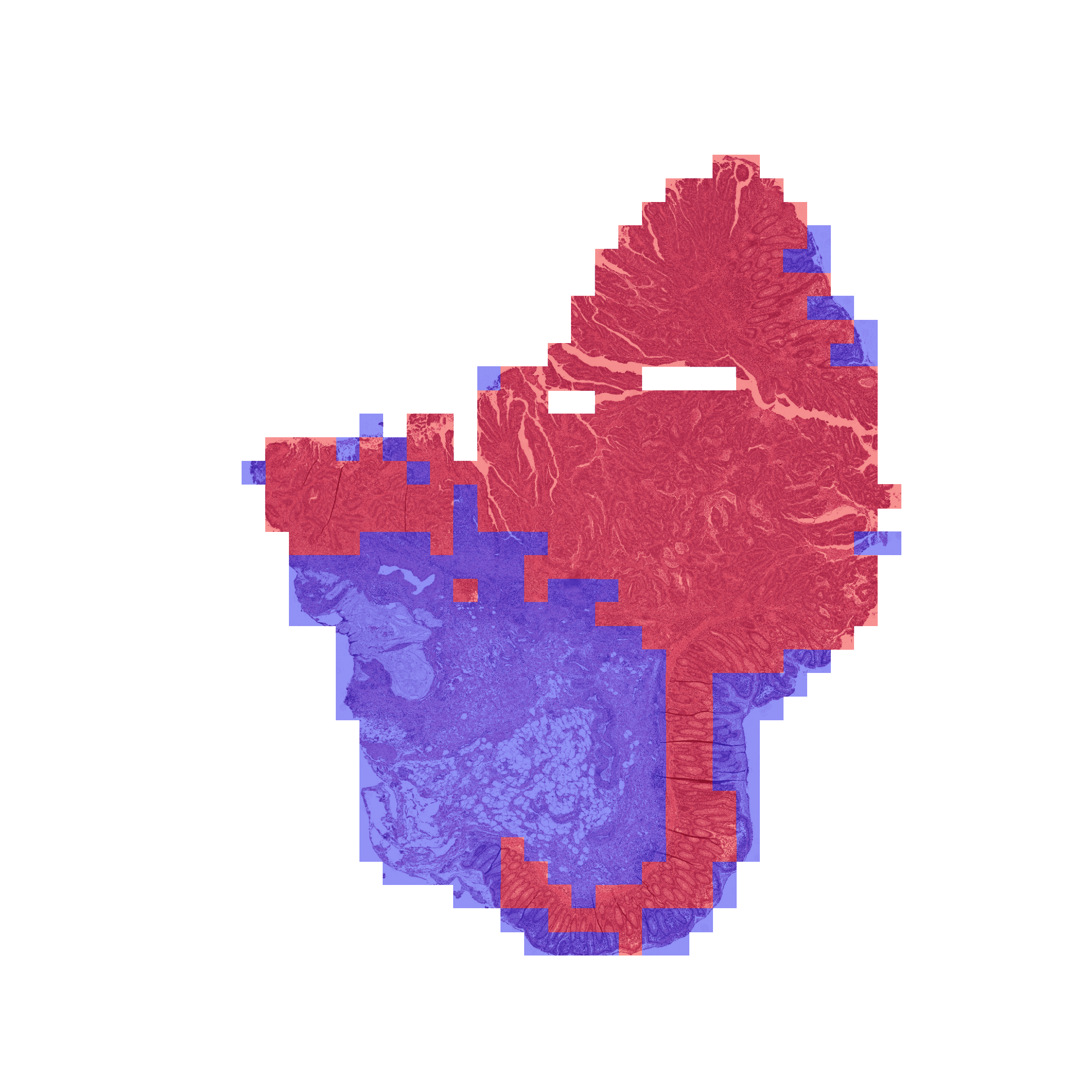} &
\inc{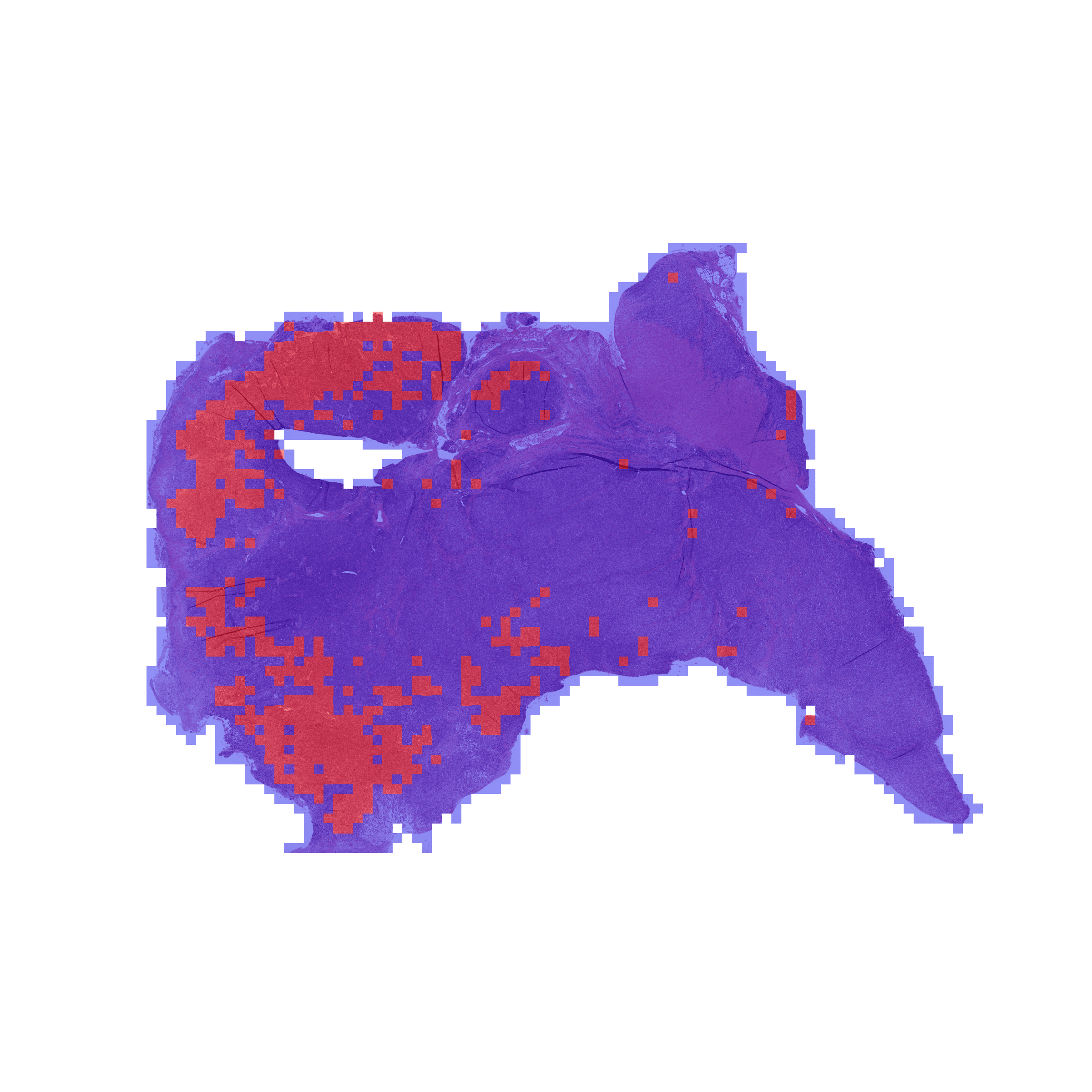} &
\inc{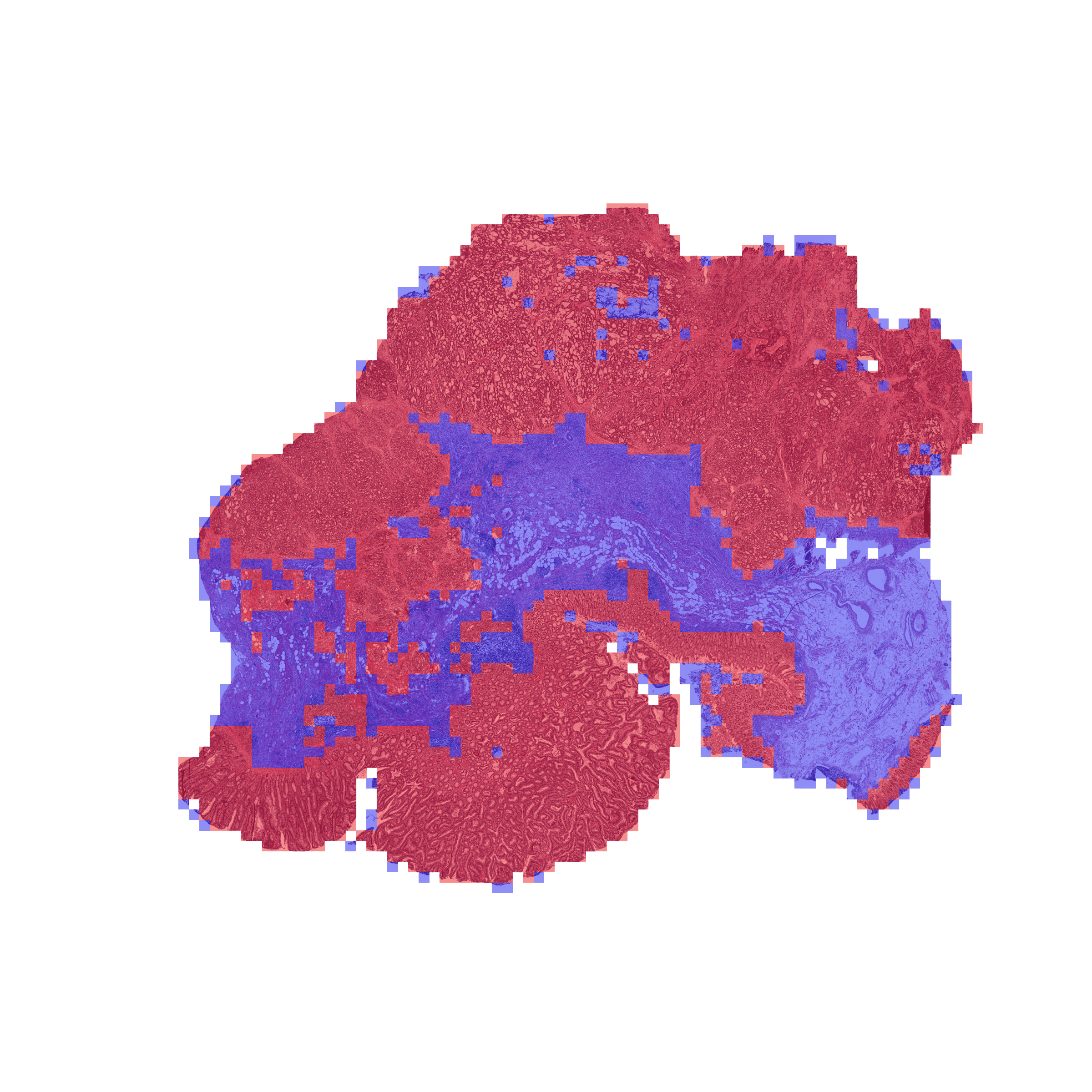} &
\multicolumn{2}{c}{\incc{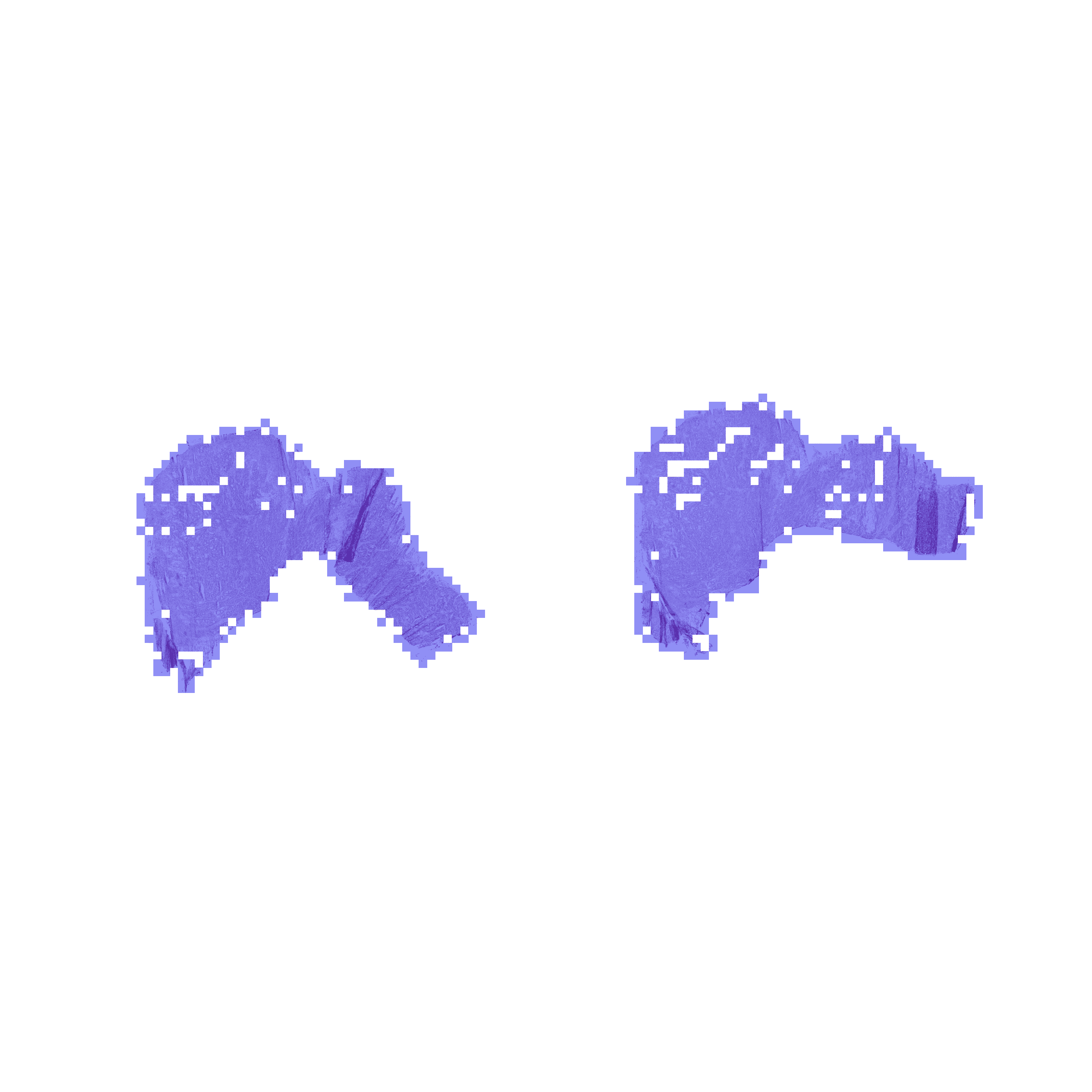}}
\\[\himg]
& F1=0.84 & F1=0.48 & F1=0.75 & F1=0.49 \\[\htext]
Best local model  &
\inc{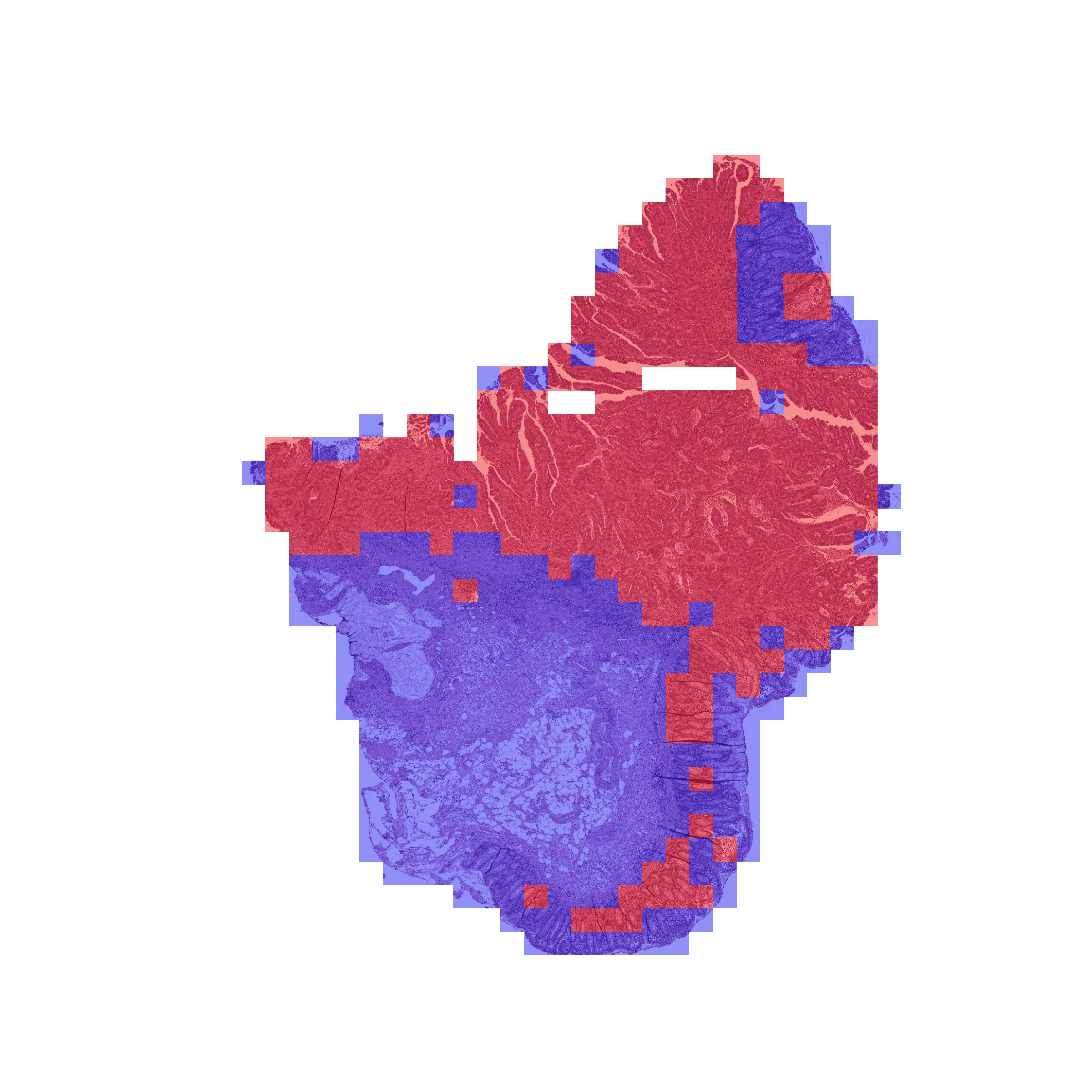} &
\inc{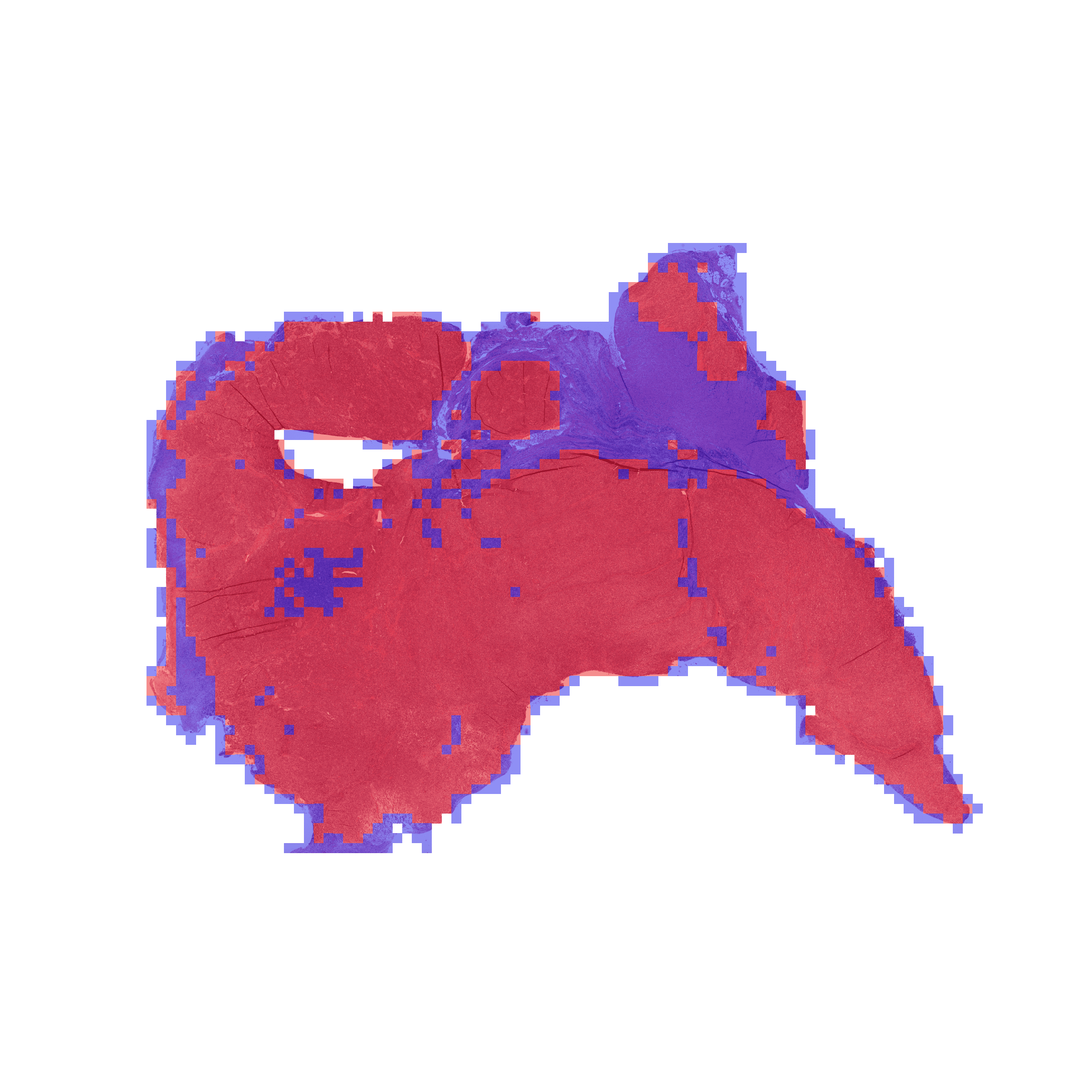} &
\inc{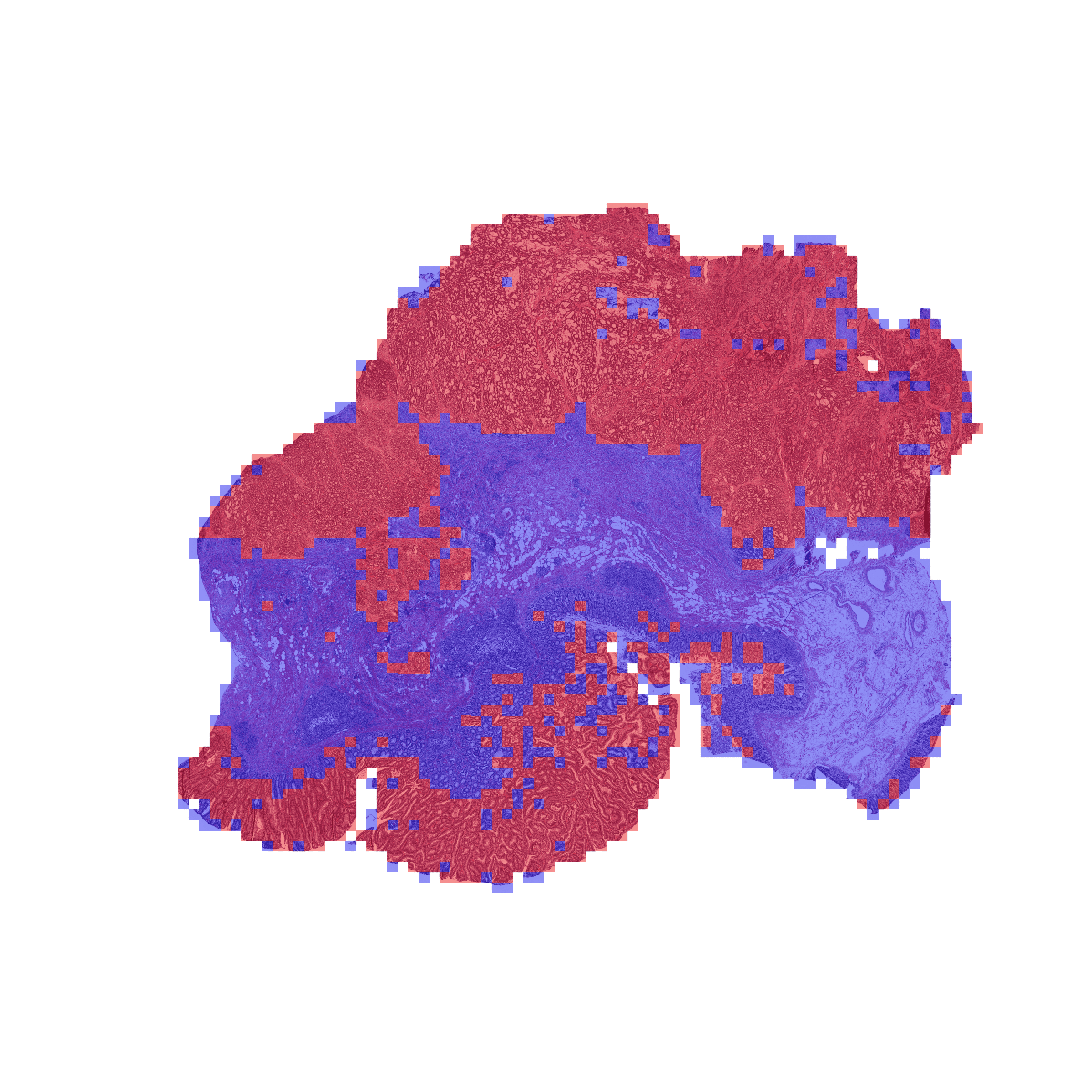} &
\incc{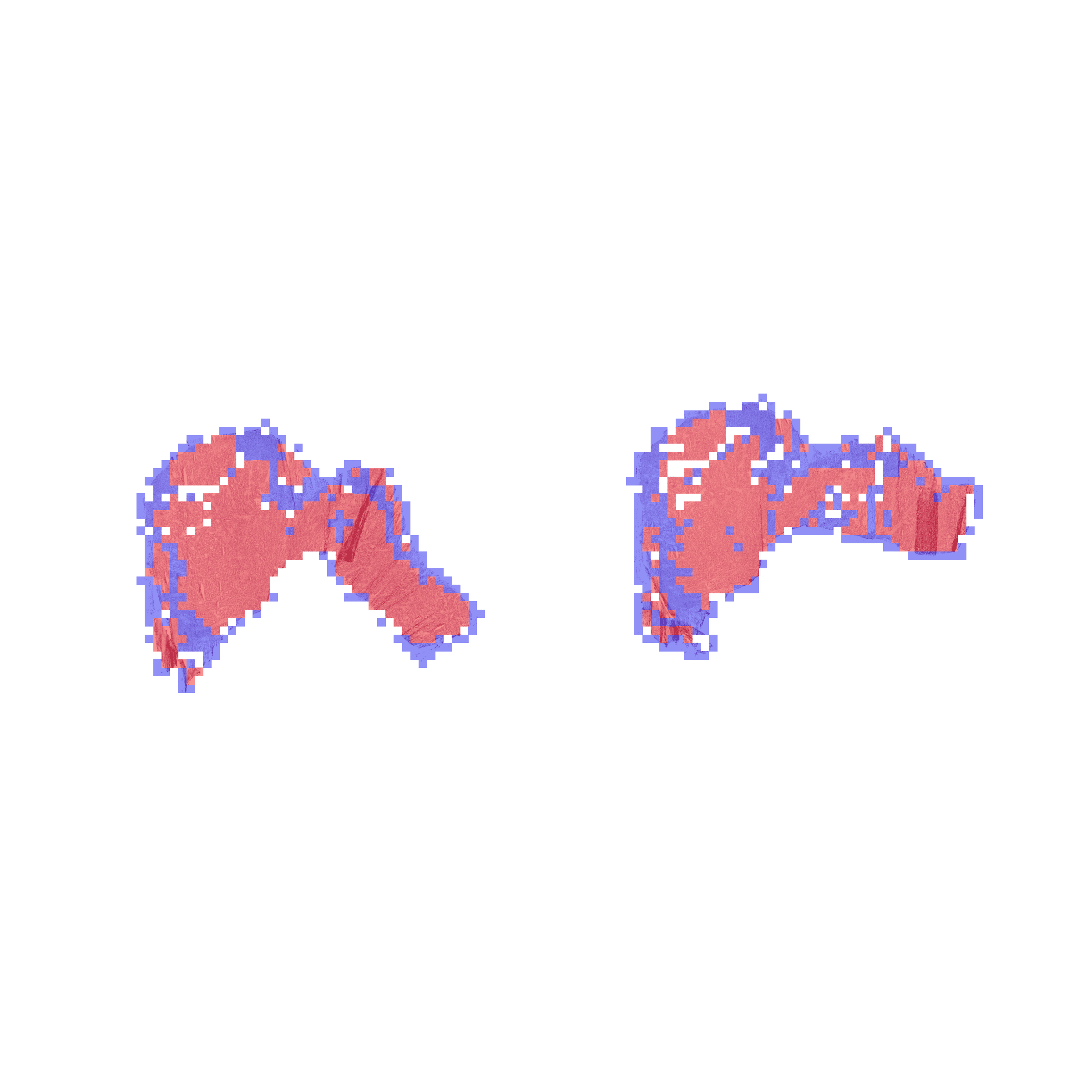} \\[\himg]
& F1=0.88 (A6) & F1=0.83 (G4) & F1=0.88 (CM) & F1=0.74 (G4) \\ 
\end{tabular}
}
\caption{Example qualitative results of different methods on WSI level. \textit{First row}: regions of interest (ROI) in four WSIs from different independent centers (respectively: AD, F4, NH, D5; original slide names are written on top of the images). \textit{Second row}: ground-truth segmentation masks. \textit{Other rows}: binary segmentation masks of different methods overlaid on images, red areas indicating ``tumor", blue areas indicating ``nontumor" predictions. Respective slide-level F1 metrics are given below each result. For the last row, best local models are selected based on comparison of all of the locally trained models by the slide-level F1 metrics; shown in parantheses are the center IDs of selected local models. All of the WSIs are from independent test sets and the presented methods are not trained on the datasets of the centers these WSIs belong to.   
}
\label{fig:segmentation-examples}
\end{figure*}


\section{Discussion and Future Directions}

In this study, we demonstrated \textit{FedDropoutAvg} method as a better way to train models with federated learning for histopathological image analysis tasks. In the application we presented, \textit{FedDropoutAvg} achieved closer performance to the conventional training where all of the data is centralized in a data lake, compared to other major federated learning approaches. We think that our strategy will allow us to achieve the goal of training better and more robust models with higher clinical usefulness while maintaining the privacy of the data via federated learning.

The dropout method in the literature~\cite{srivastava2014dropout} is a generic technique to reduce over-fitting while training a routine neural network and it is different than our approach in several ways. Firstly, it is to be used in a single neural network training and not for FL settings. Our approach is proposed for adapting model aggregation step of FL. Also, in this method only hidden layer model weight parameters are dropped-out at training time. In our proposed method, all of the model parameters might be dropped-out before federated aggregation step, including the ones in the first and final layers, and also the bias parameters. Also, the obtained aggregated global model is used both for further training rounds and/or testing. 

There have been other FL studies inspired by the drop-out method~\cite{caldas2018expanding,bouacida2020adaptive}. Namely, “Federated Dropout”~\cite{caldas2018expanding} and “Adaptive Federated Dropout”~\cite{bouacida2020adaptive}. The main goal of both of these approaches is to increase communication efficiency by decreasing the model size to be sent and received by the local clients(mobile devices). Since those studies are for FL in mobile devices, communication efficiency is an important aspect due to the large number of devices involved in the training.  In these approaches, each local client trains a smaller model (a submodel of the global model) to reduce model size and update size, while the server has the whole global model. Clients train a selected subset of the global model, and it is either a random subset~\cite{caldas2018expanding} or dynamically selected subset~\cite{bouacida2020adaptive}. Then, the server maps and reunites those smaller locally trained models into the global model. Differently than these approaches, we propose to train the same whole global model architecture locally at each client, achieving better trained local models and a much more flexible drop-out application at aggregation time.

Although there are other FL optimization methods in the literature, those are usually different additions to the FedAvg method. Because of this, in this study, we only used the major ones for comparison. We strongly believe that our proposed approach can easily be combined with other additional optimization techniques (e.g., gradient averaging with momentum of~\cite{remedios2020federated}).

As an added benefit, our proposed method could provide gains in both communication efficiency and the total amount of computation time for a real-world histopathology image analysis system. For example, when $cdr=0.2$, only 8 random clients participate at each round out of 11 clients. As a result, at each round, computation will only take place at the selected subset of clients. The clients who are not selected do not need to communicate with the server at the end of the round, providing a straightforward way to increase communication efficiency. Also, total amount of computation time will be decreased, since it is proportional to total amount of data of the selected clients. We did not provide an analysis of this aspect since the amount of data of each local training client is very heterogeneous in our dataset. In the future, additional multiple experiments can be done to understand computational efficiency of the proposed method.

We acknowledge that due to the client (center) drop out mechanism of the \textit{FedDropoutAvg}, it might be difficult or even impossible to deploy this method for a setting with a small number of participating centers, which is usually the case for medical image analysis studies. However, we believe that \textit{FedDropoutAvg} could significantly improve the performance of the final models of future studies which have high number of centers participating.

\section{CONCLUSIONS}
\label{sec:conclusion}
Federated learning can help different institutions contribute to the training of powerful models without requiring any training data to be shared. In this paper, we proposed \textit{FedDropoutAvg} and explored this federated training approach for real-world multi-site histopathology image classification and compared it with various existing federated learning methods. We evaluated the trained models on an independent test set of clients which have not participated in the training process. We showed that by using the proposed federated learning method, it is possible to achieve a classification performance comparable to a centralized model that requires data from all the clients used for training the model. In this study, we did not examine the privacy limitations of the proposed approach. We did not consider the data leakage from the model parameters if someone attempts to reconstruct the data using the model parameters exchanged during the federated training (i.e., a model inversion attack). In future, the effects of the proposed approach on privacy and combination of the proposed approach with different privacy-preserving techniques can be examined. Model aggregation is a critical piece of the FL paradigm and the improvement in performance and generalization ability of this new federated aggregation method has rich potential for usage in the future FL models. 

\section*{Acknowledgments}
This work was partly supported by the Innovate UK grant (18181) for PathLAKE project.

This research study was conducted retrospectively using human subject data made available in open access by TCGA Research Network \\ (https://www.cancer.gov/tcga). Ethical approval was not required as confirmed by the license attached with the open access data.

\bibliographystyle{IEEEtran}
\bibliography{main} 

\begin{thebibliography}{10}
\providecommand{\url}[1]{#1}
\csname url@samestyle\endcsname
\providecommand{\newblock}{\relax}
\providecommand{\bibinfo}[2]{#2}
\providecommand{\BIBentrySTDinterwordspacing}{\spaceskip=0pt\relax}
\providecommand{\BIBentryALTinterwordstretchfactor}{4}
\providecommand{\BIBentryALTinterwordspacing}{\spaceskip=\fontdimen2\font plus
\BIBentryALTinterwordstretchfactor\fontdimen3\font minus
  \fontdimen4\font\relax}
\providecommand{\BIBforeignlanguage}[2]{{%
\expandafter\ifx\csname l@#1\endcsname\relax
\typeout{** WARNING: IEEEtran.bst: No hyphenation pattern has been}%
\typeout{** loaded for the language `#1'. Using the pattern for}%
\typeout{** the default language instead.}%
\else
\language=\csname l@#1\endcsname
\fi
#2}}
\providecommand{\BIBdecl}{\relax}
\BIBdecl

\bibitem{litjens2017survey}
G.~Litjens, T.~Kooi, B.~E. Bejnordi, A.~A.~A. Setio, F.~Ciompi, M.~Ghafoorian,
  J.~A. Van Der~Laak, B.~Van~Ginneken, and C.~I. S{\'a}nchez, ``A survey on
  deep learning in medical image analysis,'' \emph{Medical image analysis},
  vol.~42, pp. 60--88, 2017.

\bibitem{konevcny2016federated}
J.~Kone{\v{c}}n{\`y}, H.~B. McMahan, D.~Ramage, and P.~Richt{\'a}rik,
  ``Federated optimization: Distributed machine learning for on-device
  intelligence,'' \emph{arXiv preprint arXiv:1610.02527}, 2016.

\bibitem{mcmahan2017communication}
B.~McMahan, E.~Moore, D.~Ramage, S.~Hampson, and B.~A. y~Arcas,
  ``Communication-efficient learning of deep networks from decentralized
  data,'' in \emph{Artificial Intelligence and Statistics}.\hskip 1em plus
  0.5em minus 0.4em\relax PMLR, 2017, pp. 1273--1282.

\bibitem{kaissis2020secure}
G.~A. Kaissis, M.~R. Makowski, D.~R{\"u}ckert, and R.~F. Braren, ``Secure,
  privacy-preserving and federated machine learning in medical imaging,''
  \emph{Nature Machine Intelligence}, vol.~2, no.~6, pp. 305--311, 2020.

\bibitem{li2018federated}
T.~Li, A.~K. Sahu, M.~Zaheer, M.~Sanjabi, A.~Talwalkar, and V.~Smith,
  ``Federated optimization in heterogeneous networks,'' \emph{arXiv preprint
  arXiv:1812.06127}, 2018.

\bibitem{li2019privacy}
W.~Li, F.~Milletar{\`\i}, D.~Xu, N.~Rieke, J.~Hancox, W.~Zhu, M.~Baust,
  Y.~Cheng, S.~Ourselin, M.~J. Cardoso \emph{et~al.}, ``Privacy-preserving
  federated brain tumour segmentation,'' in \emph{International Workshop on
  Machine Learning in Medical Imaging}.\hskip 1em plus 0.5em minus 0.4em\relax
  Springer, 2019, pp. 133--141.

\bibitem{sheller2018multi}
M.~J. Sheller, G.~A. Reina, B.~Edwards, J.~Martin, and S.~Bakas,
  ``Multi-institutional deep learning modeling without sharing patient data: A
  feasibility study on brain tumor segmentation,'' in \emph{International
  MICCAI Brainlesion Workshop}.\hskip 1em plus 0.5em minus 0.4em\relax
  Springer, 2018, pp. 92--104.

\bibitem{roy2019braintorrent}
A.~G. Roy, S.~Siddiqui, S.~P{\"o}lsterl, N.~Navab, and C.~Wachinger,
  ``Braintorrent: A peer-to-peer environment for decentralized federated
  learning,'' \emph{arXiv preprint arXiv:1905.06731}, 2019.

\bibitem{li2020multi}
X.~Li, Y.~Gu, N.~Dvornek, L.~H. Staib, P.~Ventola, and J.~S. Duncan,
  ``Multi-site fmri analysis using privacy-preserving federated learning and
  domain adaptation: Abide results,'' \emph{Medical Image Analysis}, vol.~65,
  p. 101765, 2020.

\bibitem{silva2020fed}
S.~Silva, A.~Altmann, B.~Gutman, and M.~Lorenzi, ``Fed-biomed: A general
  open-source frontend framework for federated learning in healthcare,'' in
  \emph{Domain Adaptation and Representation Transfer, and Distributed and
  Collaborative Learning}.\hskip 1em plus 0.5em minus 0.4em\relax Springer,
  2020, pp. 201--210.

\bibitem{sarhan2020fairness}
M.~H. Sarhan, N.~Navab, A.~Eslami, and S.~Albarqouni, ``On the fairness of
  privacy-preserving representations in medical applications,'' in \emph{Domain
  Adaptation and Representation Transfer, and Distributed and Collaborative
  Learning}.\hskip 1em plus 0.5em minus 0.4em\relax Springer, 2020, pp.
  140--149.

\bibitem{remedios2020federated}
S.~W. Remedios, J.~A. Butman, B.~A. Landman, and D.~L. Pham, ``Federated
  gradient averaging for multi-site training with momentum-based optimizers,''
  in \emph{Domain Adaptation and Representation Transfer, and Distributed and
  Collaborative Learning}.\hskip 1em plus 0.5em minus 0.4em\relax Springer,
  2020, pp. 170--180.

\bibitem{roth2020federated}
H.~R. Roth, K.~Chang, P.~Singh, N.~Neumark, W.~Li, V.~Gupta, S.~Gupta, L.~Qu,
  A.~Ihsani, B.~C. Bizzo \emph{et~al.}, ``Federated learning for breast density
  classification: A real-world implementation,'' in \emph{Domain Adaptation and
  Representation Transfer, and Distributed and Collaborative Learning}.\hskip
  1em plus 0.5em minus 0.4em\relax Springer, 2020, pp. 181--191.

\bibitem{wang2020automated}
P.~Wang, C.~Shen, H.~R. Roth, D.~Yang, D.~Xu, M.~Oda, K.~Misawa, P.-T. Chen,
  K.-L. Liu, W.-C. Liao \emph{et~al.}, ``Automated pancreas segmentation using
  multi-institutional collaborative deep learning,'' in \emph{Domain Adaptation
  and Representation Transfer, and Distributed and Collaborative
  Learning}.\hskip 1em plus 0.5em minus 0.4em\relax Springer, 2020, pp.
  192--200.

\bibitem{lu2020federated}
M.~Y. Lu, D.~Kong, J.~Lipkova, R.~J. Chen, R.~Singh, D.~F. Williamsona, T.~Y.
  Chena, and F.~Mahmood, ``Federated learning for computational pathology on
  gigapixel whole slide images,'' \emph{arXiv preprint arXiv:2009.10190}, 2020.

\bibitem{andreux2020siloed}
M.~Andreux, J.~O. du~Terrail, C.~Beguier, and E.~W. Tramel, ``Siloed federated
  learning for multi-centric histopathology datasets,'' in \emph{Domain
  Adaptation and Representation Transfer, and Distributed and Collaborative
  Learning}.\hskip 1em plus 0.5em minus 0.4em\relax Springer, 2020, pp.
  129--139.

\bibitem{li2019convergence}
X.~Li, K.~Huang, W.~Yang, S.~Wang, and Z.~Zhang, ``On the convergence of fedavg
  on non-iid data,'' \emph{arXiv preprint arXiv:1907.02189}, 2019.

\bibitem{bonawitz2019towards}
K.~Bonawitz, H.~Eichner, W.~Grieskamp, D.~Huba, A.~Ingerman, V.~Ivanov,
  C.~Kiddon, J.~Kone{\v{c}}n{\`y}, S.~Mazzocchi, H.~B. McMahan \emph{et~al.},
  ``Towards federated learning at scale: System design,'' \emph{arXiv preprint
  arXiv:1902.01046}, 2019.

\bibitem{li2020federated}
T.~Li, A.~K. Sahu, A.~Talwalkar, and V.~Smith, ``Federated learning:
  Challenges, methods, and future directions,'' \emph{IEEE Signal Processing
  Magazine}, vol.~37, no.~3, pp. 50--60, 2020.

\bibitem{srivastava2014dropout}
N.~Srivastava, G.~Hinton, A.~Krizhevsky, I.~Sutskever, and R.~Salakhutdinov,
  ``Dropout: a simple way to prevent neural networks from overfitting,''
  \emph{The journal of machine learning research}, vol.~15, no.~1, pp.
  1929--1958, 2014.

\bibitem{1979:otsu}
\BIBentryALTinterwordspacing
N.~Otsu, ``{A} {T}hreshold {S}election {M}ethod from {G}ray-level
  {H}istograms,'' \emph{IEEE Trans. on Systems, Man and Cybernetics}, vol.~9,
  no.~1, pp. 62--66, 1979. [Online]. Available:
  \url{http://dx.doi.org/10.1109/TSMC.1979.4310076}
\BIBentrySTDinterwordspacing

\bibitem{ResNet}
K.~He, X.~Zhang, S.~Ren, and J.~Sun, ``Deep residual learning for image
  recognition,'' in \emph{The IEEE Conf. on Computer Vision and Pattern
  Recognition (CVPR)}, June 2016, pp. 770--778.

\bibitem{kather-msi}
J.~N. Kather, A.~T. Pearson, N.~Halama, D.~J{\"a}ger, J.~Krause, S.~H. Loosen,
  A.~Marx, P.~Boor, F.~Tacke, U.~P. Neumann \emph{et~al.}, ``Deep learning can
  predict microsatellite instability directly from histology in
  gastrointestinal cancer,'' \emph{Nature Medicine}, vol.~25, no.~7, pp.
  1054--1056, 2019.

\bibitem{TIA-crc_grade}
\BIBentryALTinterwordspacing
M.~Shaban, R.~Awan, M.~M. Fraz, A.~Azam, Y.-W. Tsang, D.~Snead, and N.~M.
  Rajpoot, ``Context-aware convolutional neural network for grading of
  colorectal cancer histology images,'' \emph{IEEE Trans. on Med. Imag.}, pp.
  1--1, 2020. [Online]. Available:
  \url{https://warwick.ac.uk/fac/sci/dcs/research/tia/data/extended_crc_grading/}
\BIBentrySTDinterwordspacing

\bibitem{bilal2021novel}
M.~Bilal, S.~E.~A. Raza, A.~Azam, S.~Graham, M.~Ilyas, I.~A. Cree, D.~Snead,
  F.~Minhas, and N.~M. Rajpoot, ``Development and validation of a weakly
  supervised deep learning framework to predict the status of molecular
  pathways and key mutations in colorectal cancer from routine histology
  images: a retrospective study,'' \emph{The Lancet Digital Health}, 2021.

\bibitem{GN}
Y.~Wu and K.~He, ``Group normalization,'' in \emph{Proc. of the European Conf.
  on Computer Vision (ECCV)}, 2018, pp. 3--19.

\bibitem{hsieh2020non}
K.~Hsieh, A.~Phanishayee, O.~Mutlu, and P.~Gibbons, ``The non-iid data quagmire
  of decentralized machine learning,'' in \emph{International Conference on
  Machine Learning}.\hskip 1em plus 0.5em minus 0.4em\relax PMLR, 2020, pp.
  4387--4398.

\bibitem{caldas2018expanding}
S.~Caldas, J.~Kone{\v{c}}ny, H.~B. McMahan, and A.~Talwalkar, ``Expanding the
  reach of federated learning by reducing client resource requirements,''
  \emph{arXiv preprint arXiv:1812.07210}, 2018.

\bibitem{bouacida2020adaptive}
N.~Bouacida, J.~Hou, H.~Zang, and X.~Liu, ``Adaptive federated dropout:
  Improving communication efficiency and generalization for federated
  learning,'' \emph{arXiv preprint arXiv:2011.04050}, 2020.

\end{thebibliography}

\end{document}